\documentclass[10pt,twocolumn,letterpaper]{article}

\usepackage[pagenumbers]{cvpr} % To force page numbers, e.g. for an arXiv version

\usepackage[dvipsnames]{xcolor}

\usepackage[accsupp]{axessibility} % Improves PDF readability for those with visual impairments.
\usepackage{graphicx}
\usepackage{amsmath}
\usepackage{amssymb}
\usepackage{booktabs}
\usepackage{comment}

\graphicspath{ {images/} }
\usepackage{soul}
\usepackage{xcolor}
\usepackage{xspace}
\usepackage[super]{nth}

\usepackage{pifont}% http://ctan.org/pkg/pifont

\newcommand{\mypar}[1]{\vspace{1mm}\noindent{\bf #1}}

\definecolor{asparagus}{rgb}{0.53, 0.66, 0.42}
\definecolor{armygreen}{rgb}{0.29, 0.33, 0.13}
\definecolor{awesome}{rgb}{1.0, 0.13, 0.32}
\definecolor{applegreen}{rgb}{0.55, 0.71, 0.0}

\newcommand{\mtd}{M\&M VTO\xspace}

\definecolor{cvprblue}{rgb}{0.21,0.49,0.74}
\usepackage[pagebackref,breaklinks,colorlinks,citecolor=cvprblue]{hyperref}

\def\paperID{4593} % *** Enter the Paper ID here
\def\confName{CVPR}
\def\confYear{2024}

\title{{\mtd}: Multi-Garment Virtual Try-On and Editing}

\author{
Luyang Zhu\textsuperscript{1,2}\footnotemark\qquad
Yingwei Li\textsuperscript{1}\qquad
Nan Liu\textsuperscript{1}\qquad
Hao Peng\textsuperscript{1}\qquad \\
Dawei Yang\textsuperscript{1}\qquad
Ira Kemelmacher-Shlizerman\textsuperscript{1,2} \\
\textsuperscript{1}Google Research\qquad
\textsuperscript{2}University of Washington
}

\begin{document}
\twocolumn[{%
\renewcommand\twocolumn[1][]{#1}%
\maketitle
\begin{center}
    \centering
    \includegraphics[width=1\textwidth]{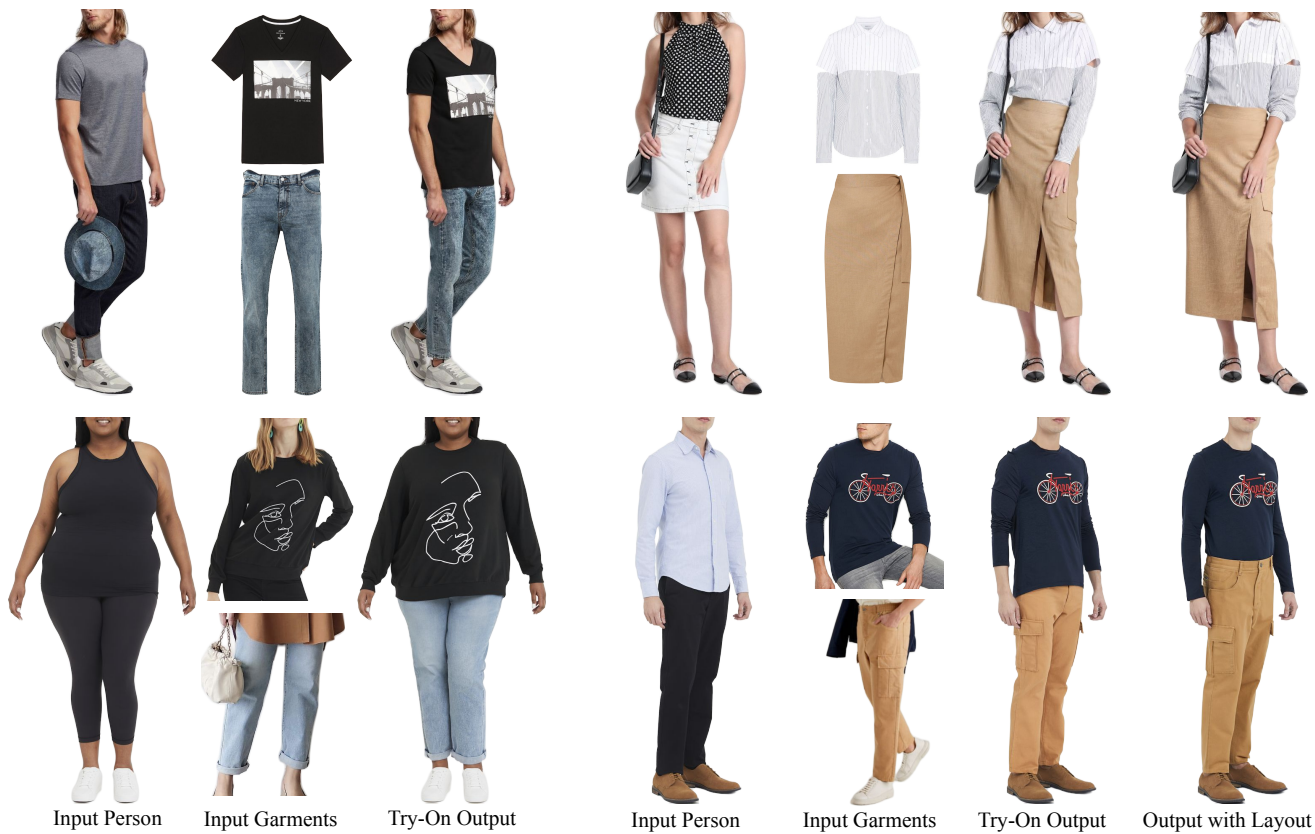}
    \vspace{-15pt}
    \captionof{figure}{\small 
Given an input person image, multiple garments, \mtd can output a virtual try-on visualization of how those garments would look on the person. Our model performs well across various body shapes, poses, and garments. In addition, it allows layout to be changed, \eg, ``roll up the sleeves'' (top rightmost column), and ``tuck in the shirt and roll down the sleeves'' (bottom rightmost column). 
    }
    \label{fig:teaser}
\end{center}%
}]

\footnotetext[1]{Work done while author was an intern at Google.}

\maketitle
\begin{abstract}

We present {\mtd}–a mix and match virtual try-on method that takes as input multiple garment images, text description for garment layout and an image of a person.  An example input includes: an image of a shirt, an image of a pair of pants, ``rolled sleeves, shirt tucked in'', and an image of a person. The output is a visualization of how those garments (in the desired layout) would look like on the given person. Key contributions of our method are: 
1)  a single stage diffusion based model, with no super resolution cascading, that allows to mix and match multiple garments at $1024\mathord\times\mathord512$ resolution preserving and warping intricate garment details,
2) architecture design (VTO UNet Diffusion Transformer) to disentangle denoising from person specific features, allowing for a highly effective finetuning strategy for identity preservation (6MB  model per individual vs 4GB achieved with, e.g., dreambooth finetuning); solving a common identity loss problem in current virtual try-on methods,
3)  layout control for multiple garments via text inputs finetuned over PaLI-3~\cite{chen2023pali} for virtual try-on task.  Experimental results indicate that {\mtd} achieves state-of-the-art performance both qualitatively and quantitatively, as well as opens up new opportunities for virtual try-on via language-guided and multi-garment try-on.

\end{abstract}

\begin{figure*}[t]
\begin{center}
  \includegraphics[width=1.0\linewidth]{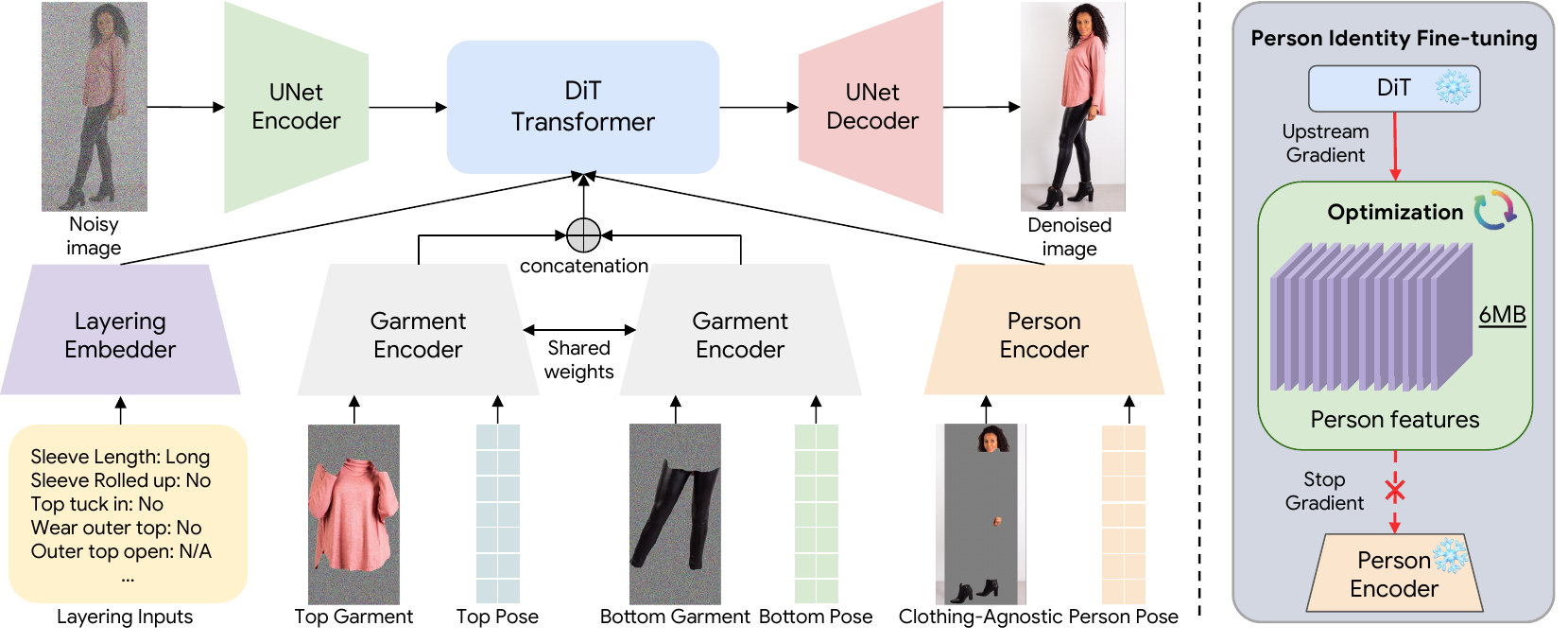}
\end{center}
\vspace{-5pt}
  \caption{\textbf{Overview of {\mtd}.}  \textbf{Left}: Given multiple garments (top and bottom in this case, full-body garment not shown for this example), layout description, and a person image, our method enables multi-garment virtual try-on. \textbf{Right}: By freezing all the parameters, we   optimize  person feature embeddings extracted from the person encoder to improve person identity for a specific input image. The fine-tuning process recovers the information lost via agnostic computation. }
\label{fig:pipeline}
\end{figure*}
\section{Introduction}
\label{sec:intro}

Virtual try-on (VTO) is the task of synthesizing how a person would look in various garments based on provided garment photos and a person photo. Ideally the synthesis is high resolution, showcasing the intricate details of garments, while at the same time representing the body shape, pose, and identity of the person accurately.  In this paper, we focus specifically on multiple garment VTO and editing. For example, a user of our method would provide one or more photos for garments, e.g., shirt, pants, and one photo of a person, with additional optional text input to request a layout, e.g., “shirt tucked out, rolled sleeves”. The garment photos could be either a product photo (layflat) or the garment as worn by a different person. The person photo would be a full field-of-view photo showing the person head to toe.  Our method, which we named, {\mtd}, outputs a visualization of how the person looks in those garments. Figure~\ref{fig:teaser} shows a couple of examples.  

Redefining the VTO problem as multiple-garment VTO, rather than the commonly targeted single garment VTO, allowed us to deeply rethink architecture design and solve several open problems in multi, as well as single VTO networks, in addition to opening up the new possibilities for mix and match and editing layouts. 

Two of the most challenging VTO problems are (1) how to preserve the small but important details of garments while warping the garment to match various body shapes, and (2) how to preserve the identity of the person without leaking the original garments that the person was wearing to the final result.  state-of-the-art methods came close for single garment VTO by leveraging the power of diffusion, and building networks that denoise while warping, e.g., Parallel-Unet~\cite{Zhu_2023_CVPR_tryondiffusion}. To address (1), however, the network requires to max out the number of parameters and a memory heavy Parallel-UNet to warp a single garment. For (2) a ``clothing-agnostic" representation is typically used for the person image to erase the current garment to be replaced by VTO, but at same time it removes a significant amount of identity information, with the network needing to hallucinate the rest, resulting in loss of characteristics like tattoos, body shape or muscle information.  

With more garments, as in multi-garment VTO, the number of pixels needed to go through the network triples, so the same number of parameters would create a lower quality VTO. Similarly, showing head to toe person and allowing multiple garments, means `clothing-agnostic’ representation leaves even less of the identity of the person–if just a shirt needs to be replaced, the network can still see how the bottom part of that person looks like (and shape of the legs), while if all garments are changing the agnostic would preserve even less information about the person. 

Our solution, {\mtd}, is three-fold as depicted in Figure~\ref{fig:pipeline}. First, we designed a single-stage diffusion model to directly synthesize $1024\mathord\times\mathord512$ images with no need for extra super-resolution(SR) stages as commonly done by state-of-the-art image generation techniques. We found that as we expand the scope of VTO, having cascaded design is detrimental as the base model’s low resolution assumes excessive downsampling of ground truth during training, thus losing forever garment details; as SR models depend heavily on the base model, if the details disappear they can not be upsampled effectively. Training a single stage base model just on higher resolution data, however, does not solve the problem, as the model doesn’t converge even with ideas proposed in \cite{hoogeboom2023simple,chen2023importance}. Instead we designed a progressive training strategy where model training begins with lower-resolution images and gradually moves to higher-resolution ones during the single stage training. Such a design naturally benefits training at higher resolutions by utilizing the prior learned at lower resolutions, allowing the model to better learn and refine high-frequency details. 

Second, to solve the identity loss (and/or clothing leakage) during the `clothing-agnostic' process, we propose a space saving finetuning strategy.  Rather than finetuning the entire model during post processing, as commonly done by techniques like DreamBooth \cite{ruiz2022dreambooth}, we choose to finetune person features only. We designed a VTO UNet Diffusion Transformer (VTO-UDiT) to isolate encoding of person features from the denoising process. In addition to producing much higher quality results, this design also drastically reduces finetuned model size per new individual, going from $4$GB to $6$MB. 

Third, we created text based labels representing various garment layout attributes, e.g., rolled sleeves, tucked in shirt, and open jacket. We formulated attribute extraction as an image captioning task and finetuned a PaLI-3 model ~\cite{chen2023pali} using only $1.5k$ labeled images. This allows us to automatically extract accurate labels for the whole training set. 

Above three design choices are critical in producing high quality VTO results for multi-garment scenarios. We perform detailed ablation studies, and comparisons to state-of-the-art papers to illustrate each design choice. Our method significantly outperforms others. The user study shows that our method is chosen as best ${78.5\%}$ of the time compared to state-of-the-art on multiple-garment VTO task.

\section{Related Work}
\label{sec:related_work}

In this section we will focus on related work relevant to our three key design choices described above. For a  comprehensive list of recent papers in virtual try-on we also invite the reader to review this list\footnote{https://github.com/minar09/awesome-virtual-try-on}.

\mypar{Image-Based Virtual Try-On.} The seminal VITON method~\cite{han2018viton}  proposed a warping model that estimates pixel displacements between the original garment image and target warp. Based on those displacements, it warped the garment, and then used a blending model to combine the warped garment with the person image, showing one of the first promising results for VTO. Many works followed, to improve pixel displacement estimation.   \cite{wang2018toward} proposed thin plate splines, \cite{yu2019vtnfp} predicted target segmentation and parsing for improved warping,  student-teacher approach and distillation were proposed by ~\cite{issenhuth2020not,ge2021parser}. Other efforts include adaptive parsing and second order constraint on thin plate splines~\cite{yang2020towards}, optimization to remove misalignments~\cite{choi2021viton}, leveraging dance videos to improve warping~\cite{dong2022dressing}, regularizing~\cite{Yang_2022_CVPR}, and using self and cross attention to improve flow computation~\cite{bai2022single}. With the rise of StyleGAN,~\cite{He_2022_CVPR} proposed  StyleGAN for optical flow, \cite{lee2022hrviton} proposed a generator-discriminator approach,  \cite{Yan_2023_CVPR,li2023virtual,xie2023gp} reported improved results for flow compute and inpainting by  utilization of landmarks, and \cite{chen2023size} incorporated size information.

While results were improving, there was an inherent difficulty in warping garments \textit{explicitly--pixel wise}, as there is too much variation in folds, logos, texture where a garment image needs to warp to a new  body shape. Rather than estimating flows directly, \cite{lewis2021tryongan} proposed to interpolate StyleGAN coefficients to create try-on, still lacking complex textures, though, due to the averaging nature of StyleGAN.   TryOnDiffusion \cite{Zhu_2023_CVPR_tryondiffusion} introduced a diffusion-based \cite{sohl2015deep,ho2020denoising, song2019generative}   Parallel-UNet enabling implicit  warping and blending in the same  model via cross-attention, showing significantly better results. Key limitations of that approach were incomplete garment details due to base model being only  
 $128\mathord\times\mathord128$ resolution, and  identity preservation. Finally, most of those methods are focused on single garment try-on only. 
 
\begin{figure*}
   \includegraphics[width=1.0\linewidth]{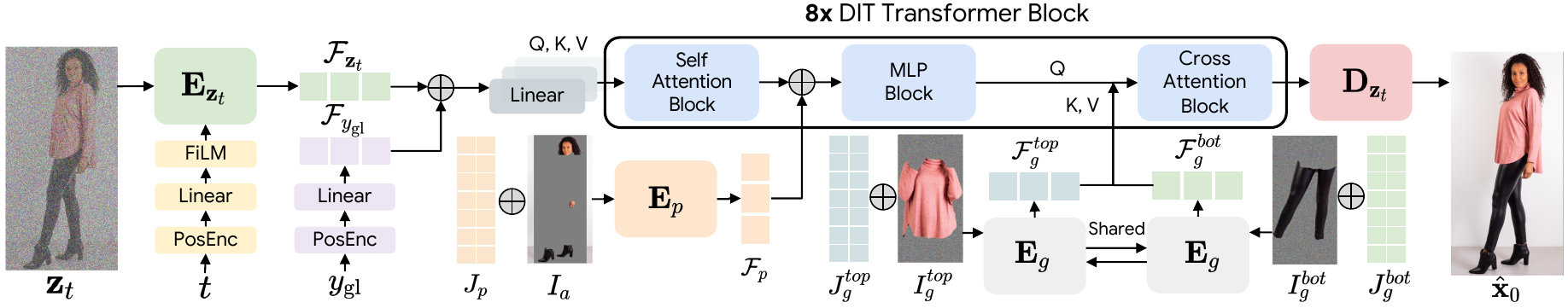}
\caption{\textbf{VTO-UDiT architecture.} For image inputs, UNet encoders ($\mathbf{E}_{\mathbf{z}_t}$, $\mathbf{E}_{p}$, $\mathbf{E}_{g}$) extract features maps ($\mathcal{F}_{\mathbf{z}_t}$, $\mathcal{F}_{p}$, $\mathcal{F}_{g}^{\kappa}$) from $\mathbf{z}_t$, $I_{a}$, $I_{c}^{\kappa}$, respectively, with $\kappa \in \{\text{upper}, \text{lower}, \text{full}\}$. Diffusion timestep $t$ and garment attributes $y_{\text{gl}}$ are embedded with sinusoidal positional encoding, followed by a linear layer. The embeddings ($\mathcal{F}_{t}$ and $\mathcal{F}_{y_{\text{gl}}}$) are then used to modulate features with FiLM \cite{dumoulin2018feature} or concatenated to the key-value feature of self-attention in DiT similar to \cite{saharia2022photorealistic}. Following \cite{Zhu_2023_CVPR_tryondiffusion}, spatially aligned features($\mathcal{F}_{\mathbf{z}_t}$, $\mathcal{F}_{p}$) are concatenated whereas $\mathcal{F}_{g}^{\kappa}$ are implicitly warped with cross-attention blocks. The final denoised image $\hat{\mathbf{x}}_0$ is obtained with decoder $\mathbf{D}_{\mathbf{z}_t}$, which is architecturally symmetrical to $\mathbf{E}_{\mathbf{z}_t}$.}
\label{fig:vtoudit}
\end{figure*}
\mypar{Finetuning Diffusion.} 
As finetuning is a general concept and wasn't much used for VTO, we will review recent works for any general finetuning. Sometimes also called personalization \cite{gal2022image}, finetuning is the task of adjusting an existing, say text to image generation model, to a specific task, e.g., style transfer. Dreambooth~\cite{ruiz2022dreambooth} showed fantastic results by finetuning on a few images, and accompanying text, to bind a unique identifier with a specific subject. ~\cite{Gal:2023aa,Li:2023aa} learned encoders to transfer visual concept into textual embeddings. \cite{alaluf2023neural} created a network that maps noise timestamp and layer to text token space. To improve multi-concept composition~\cite{liu2022compositional}, Custom Diffusion~\cite{kumari2022customdiffusion} optimized concept embeddings along with key and value projection matrices of  cross attention layers in the text-to-image model. In contrast, our approach is tailored for VTO and requires only $\mathbf{6}$MB of parameters per person during the inference phase.  

\mypar{Image Editing with Diffusion Models.}  Editing of general images with diffusion initially utilized  image masks~\cite{nichol2021glide, meng2021sdedit,lugmayr2022repaint,saharia2022palette,couairon2022diffedit,avrahami2022blended}. SDEdit~\cite{meng2021sdedit} added noise to the inputs and then subsequently denoised them through a stochastic  process.  Palette~\cite{saharia2022palette}  trained a  conditional diffusion model for specific edit tasks.  BlendedDiffusion~\cite{avrahami2022blended}, inspired by CLIP guided diffusion~\cite{clip_guided_diffusion}, utilized CLIP text encoder~\cite{radford2021learning} and spatial masks to edit images by blending noised input images with locally generated contents. Requiring masks is not applicable to VTO tasks e.g., tuck this shirt in. 

The success of text to image diffusion models ~\cite{nichol2021glide,ramesh2022hierarchical, ho2022imagen,rombach2022high} led to text-based image editing ~\cite{couairon2022diffedit,hertz2022prompt,brooks2023instructpix2pix,mokady2023null,kim2022diffusionclip,sheynin2022knn,valevski2022unitune,kawar2023imagic,wang2023imagen}. For example, DiffEdit~\cite{couairon2022diffedit} infers a region mask based on text instructions, and then guides image editing using inverted noise resulted from DDIM inversion process~\cite{song2020denoising}. Prompt-to-Prompt (P2P)~\cite{hertz2022prompt} edits images using only text by manipulating the cross-attention scores conditioned on inverted latents. 
Null-text inversion~\cite{mokady2023null} optimized on null-text embeddings by minimizing differences between latent codes from unconditional inversion process and conditional one. 
InstructPix2Pix~\cite{brooks2023instructpix2pix} directly manipulates image in the denoising process by using finetuned Stable Diffusion trained on paired examples generated using P2P technique with given editing instructions.   Generally, text based editing, while allowing for easier input (compared to masks), often creates the edit but fails to preserve original image details, e.g., in VTO case the original garment details are lost with such techniques. We solve it via VTO specific finetuning on PaLI-3 and then using it as condition in the network. 
\begin{figure*}[htb]
\begin{center}
   \includegraphics[width=1.0\linewidth]{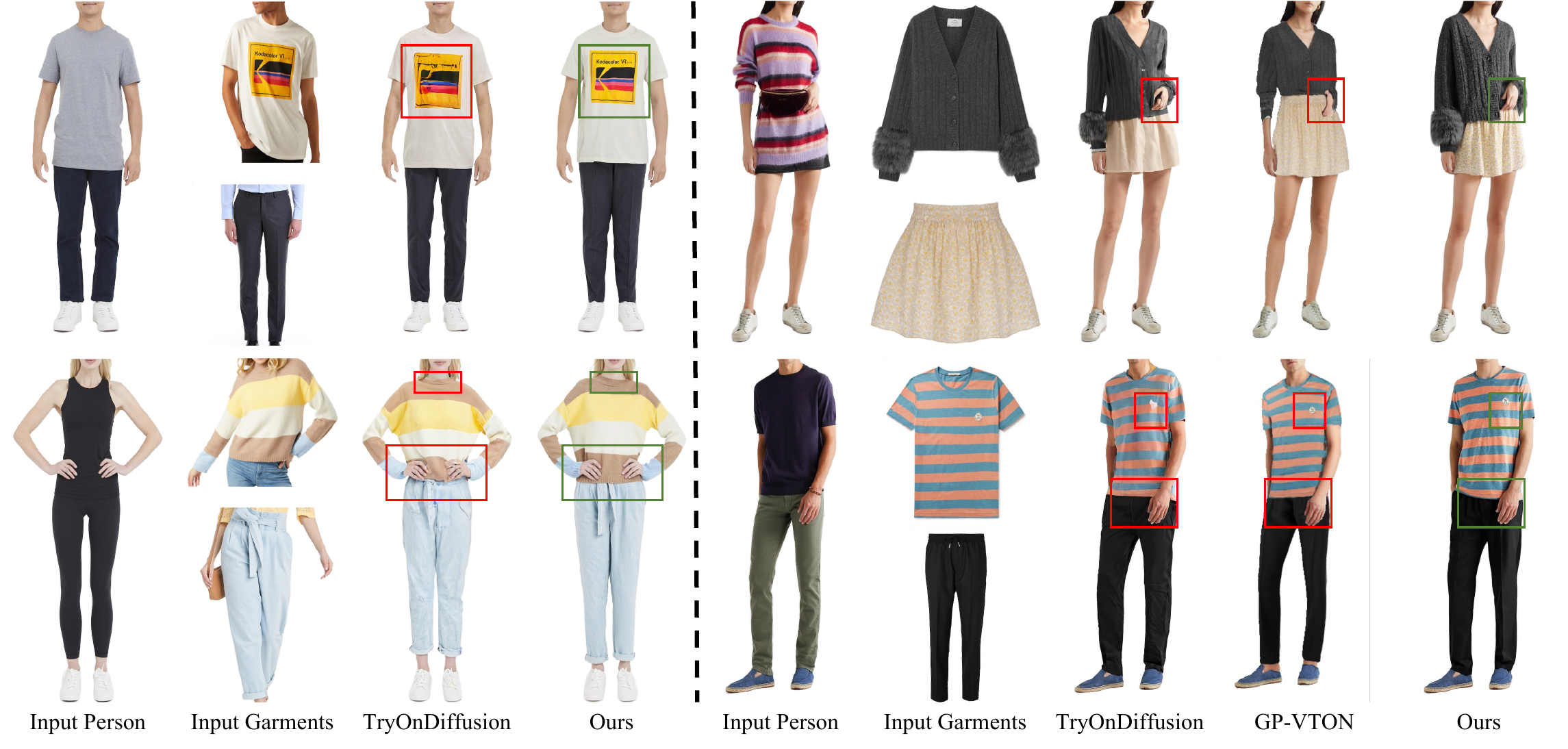}
\end{center}
\vspace{-5mm}
\caption{\textbf{Qualitative  Comparison with existing Try-On methods.} On the left, we compare with TryOnDiffusion~\cite{Zhu_2023_CVPR_tryondiffusion} on our test set and further evaluate on DressCode \cite{morelli2022dress} dataset, as shown on the right. Our method can generate better garment details and layouts.}
\label{fig:compare_tryondiffusion}
\vspace{-5mm}
\end{figure*}

\section{Method}
\label{sec:method}
 Given a person image $I_{p}$, an upper-body garment image $I_{g}^{\text{upper}}$, a lower-body garment image $I_{g}^{\text{lower}}$ and a full-body garment image $I_{g}^{\text{full}}$, our method  synthesizes  VTO result $I_{\text{tr}}$ for person $p$. Optionally, a layout attribute is provided as input as well.  We begin by describing training data and its preprocessing, and then the model design of \mtd. 

\subsection{Dataset Preparation and Preprocessing}
\label{subsec::dataset_preparation}

{\mtd} is trained on pairs--person image $I_{p}$, and a garment image $I_{g}$.  $I_{g}$ can be an image of a garment laid out on a flat surface (layflat), or an image of a person wearing the garment (most often in another pose). As the pair assumes that they share only one or two garments rather than all three of upper, lower and full, we do the following simple process. We compute a garment embedding for each of the three garments (determined by segmentation) and compare which one appears on the person image. The ones that do not are set to 0.

Each pair is then processed following \cite{Zhu_2023_CVPR_tryondiffusion}. Conditional inputs $\mathbf{c}_\text{tryon}$  includes clothing-agnostic RGB $I_{a}$, segmented garment $I_{c}^{\kappa}$, 2D pose keypoints $J_{p}$ for the person image $I_{p}$ and 2D pose keypoints $J_{g}^{\kappa}$ for garment images $I_{g}^{\kappa}$ ($J_{g}^{\kappa}$ is a vector with all -1's if $I_{g}^{\kappa}$ is a layflat garment image). To make sure that background is as tight as possible (allowing for the model to fully focus on garments) we crop and resize all  images to $1024\mathord\times\mathord512$, approximately resembling aspect ratio of a photograph of a head to toe person. 

We also introduce a layout input $y_{\text{gl}}$, defining desired attributes of the garments. We only focus on attributes that one can do in real-life, for example: roll up sleeves, tuck in the shirt, etc. rather than changing texture or garment properties.   Full set of attributes is in the supplementary material. One way to calculate attributes of each garment is by training a classifier for each attribute.  We chose instead to finetune a large vision language model (PaLI-3\cite{chen2023pali}). Specifically, we convert all attributes into a formatted text and formulate it as an image captioning task. There are two advantages for this formulation. First, vision language models have strong priors trained on large datasets and can utilize the correlation between different garment layout attributes (\eg the sleeve can not be rolled up if the sleeve type is sleeveless). Second, using a single model can also accelerate the training data generation process. Thanks to the strong prior encoded in the  PaLI-3 model, we are able to get very accurate garment attributes by finetuning PaLI-3 with only 1,500 images. To get $y_{\text{gl}}$ for each training sample, we first extract garment layout attributes relevant to the garment type $\kappa$ by running finetuned PaLI-3  on $I_{g}^{\kappa}$, and then concatenate those attributes into a single vector. Refer to the supplementary for more details.

\subsection{Single Stage {\mtd}}
\label{subsec::mixmatch_tryondiffusion}

Cascaded diffusion models, i.e., lower resolution diffusion base model, followed by super resolution models, have shown great success for text to synthetic image generation \cite{wang2023imagen,ho2022imagen}.  Similarly, for VTO \cite{Zhu_2023_CVPR_tryondiffusion} followed a similar setup where three stages were used.  For multi-garment VTO, however, such design is performing poorly, as the base model doesn't have enough capacity to create intricate warps and occlusions based on person's body shape.  We observed that high-frequency garment details are smoothed and blurred out if images are downsampled by more than 2 times. Thus, it is impossible for base diffusion models trained to preserve those garment details as their groundtruth images do not include them.

Ideally we would just  synthesize $1024\mathord\times\mathord512$ images with the base model directly.   This turned out to be a challenging task, as if the cross-attention is applied at a lower resolution, the high frequency image details are destroyed by excessive downsampling of feature maps, and the model tends to learn a global structure for the warping. On the other hand, applying cross-attention at a higher resolution does not converge under random initialization from our initial experiments.

To tackle this challenge, we use an effective progressive training paradigm for {\mtd}. The key idea is to initialize the higher resolution diffusion models using a pre-trained lower resolution one. Specifically, we first train a base diffusion model to synthesize $512\mathord\times\mathord256$ try-on results $I_{\text{tr}}^{512\mathord\times\mathord256}$, where the cross-attention happens in $32\mathord\times\mathord16$. After that, we continue to train the \textit{exact same model} to synthesize $1024\mathord\times\mathord512$ try-on results $I_{\text{tr}}^{1024\mathord\times\mathord512}$, where the cross-attention happens in $64\mathord\times\mathord32$ with the same architecture. Note that our training algorithm does not require modifying or adding new components to the architecture, all we need is to train the model with data in different resolutions, which is easy to implement. 

\begin{figure*}[htb]
\begin{center}
   \includegraphics[width=1\linewidth]{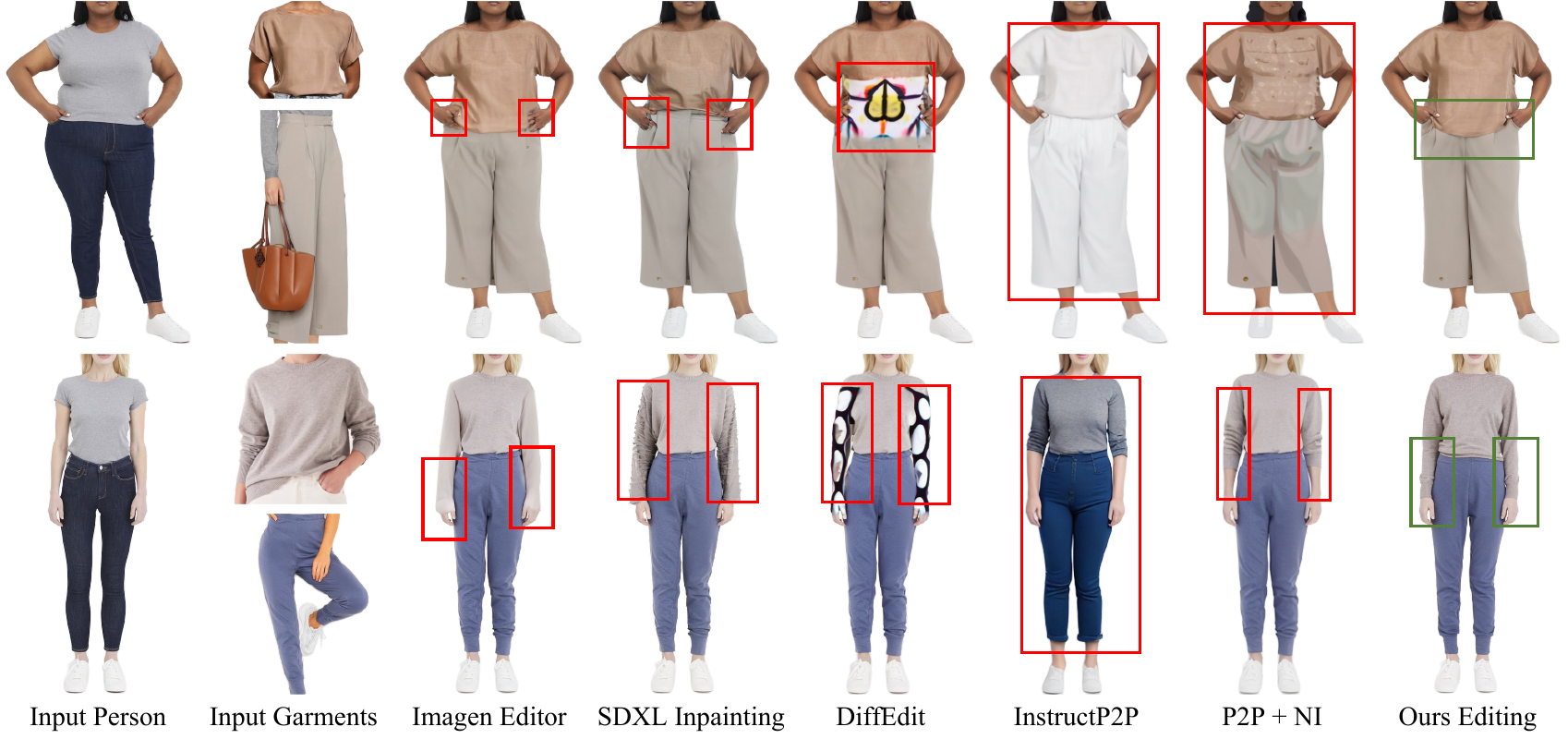}
\end{center}
\vspace{-5mm}
\caption{\textbf{Qualitative Comparison for Garment Layout Editing.} Top: editing instruction is to ``tuck out the shirt``. Bottom: ``roll down the sleeve''. Our method enables more accurate layout editing while preserving the details from the inputs. Details are provided in the Supplementary.}
\label{fig:edit}
\end{figure*}

\subsection{VTO-UDiT Architecture}
\label{subsec::tryon_udit}

The VTO-UDiT network (Figure~\ref{fig:vtoudit}) is represented as  
\begin{equation}
\label{eq:classifier-free guidance}
    \hat{\mathbf{x}}_0 = \mathbf{x}_\theta(\mathbf{z}_t, t, \mathbf{c}_\text{tryon})
\end{equation}

where $t$ is the diffusion timestep, $\mathbf{z}_t$ is the noisy image corrupted from the ground-truth $\mathbf{x}_0$ at timestep $t$, $\mathbf{c}_\text{tryon}$ is the try-on conditional inputs, and $\hat{\mathbf{x}}_0$ is the predicted clean image at timestep $t$. In practice, we follow \cite{ho2022imagen} to set the network output in $\mathbf{v}$-space to avoid color drift issues in higher resolution diffusion models. Given the predicted $\hat{\mathbf{v}}_t$, we  compute $\hat{\mathbf{x}}_0 = \alpha_{t}\mathbf{z}_t - \sigma_{t}\hat{\mathbf{v}}_t$, where $\alpha_{t}, \sigma_{t} \in (0, 1)$ control the signal-to-noise ratio.

Inspired by \cite{hoogeboom2023simple}, we change the Parallel-UNet architecture \cite{Zhu_2023_CVPR_tryondiffusion} into a UDiT architecture where the transformer block is implemented as DiT \cite{peebles2023scalable}. With the combination of UNet and DiT, the model benefits from light weight UNet as image encoders and the heavy DiT blocks to process in lower resolution feature maps for attention operations.

Moreover, the design of UDiT fully disentangle the encoding process of $\mathbf{c}_\text{tryon}$ from the denoising process, which is critical for person feature finetuning described later in Section~\ref{subsec:finetune_person}. More specifically, 1) Different UNet encoders are used to process the input images without information exchange. 2) Only $\mathbf{E}_{\mathbf{z}_t}$ takes diffusion timestep $t$ embedding as as input, while $\mathbf{E}_{p}$ and $\mathbf{E}_{g}$ do not, to fully disentangle conditional features from diffusion denoising. 3)Unlike Parallel-UNet \cite{Zhu_2023_CVPR_tryondiffusion} which updates both conditional features and noisy image features in parallel, VTO-UDiT fixes the conditional features and only updates diffusion features during the forward pass of DiT blocks. 

Also, note that all UNet encoders are fully convolutional and free of attention operations, which is preferable for progressive training mentioned in Section~\ref{subsec::mixmatch_tryondiffusion}.

\subsection{Efficient Finetuning for Person Identity}
\label{subsec:finetune_person}
A key challenge of current VTO methods is the loss of person identity due to the use of clothing-agnostic representation. To tackle this problem, we propose a space-efficient finetuning strategy based on our VTO-UDiT architecture. As described in Sec. \ref{subsec::tryon_udit},  person feature $\mathcal{F}_{p}$ is independent of diffusion or garment related features, and is kept fixed for DiT blocks where conditioning happens. Thus, we are able to directly finetune the person features instead of the whole diffusion model. This greatly reduces the optimizable weights from 4GB to 6MB. Furthermore, we found finetuning on person features will not cause the model to overfit on the particular garments worn by the target person as shown in Section~\ref{sec:experiments}.

The finetuning process needs to learn how to warp garments from varying sizes and poses on the target person, however, acquiring pairs of images of same garment and various shapes and sizes is impractical.  Instead we use pretrained {\mtd} to prepare a synthetic dataset. We segment out garments worn by the target person image, and try-on the garment on multiple person images across various poses (\eg different torso orientations and arm positions) and body shapes (from 2XS to 2XL), resulting in $150$ samples.  Since our pretrained {\mtd} can accurately preserve but warp garment details to new pose and shape, the quality of the synthetic finetuning data is high, and allows us to reconstruct the person identity  when  tested on   unseen garments.
\section{Experiments}
\label{sec:experiments}
In this section, we describe datasets, comparisons and ablations. Additional results as well as implementation details can be found in supplementary. 

\mypar{Datasets.}  Our model is trained on two types of datasets: 1) ``garment paired'' dataset of $17$ Million samples, where each sample consists of two images of the same garment in two different poses/body shapes, 2) ``layflat paired'' dataset of $1.8$ Million samples, where each sample consists of an image with garment laid out on a flat surface and an image of a person wearing the garment. For testing, we use two sets: 1)  we collected  $8,300$ triplets (top, bottom, person) that are \textit{unseen} during training, 2) we use DressCode~\cite{morelli2022dress} just for comparison with other methods that use it.

\begin{table}[t]
\centering
\scalebox{0.7}{
\begin{tabular}{ |c|ccc|ccc|  }
\hline
\multicolumn{1}{|c}{Test datasets} & \multicolumn{3}{|c}{Ours $8,300$} & \multicolumn{3}{|c|}{DressCode}\\
\hline
Methods & FID $\downarrow$ & KID $\downarrow$ & US $\uparrow$ & FID $\downarrow$ & KID $\downarrow$ & US $\uparrow$\\ 
\hline 
GP-VTON~\cite{xie2023gp} & N/A  & N/A & N/A & $38.392$ & $33.909$ & 1327 \\
TryOnDiffusion~\cite{Zhu_2023_CVPR_tryondiffusion} & $19.459$ & $17.617$ & $1526$ &  $15.944$  & $5.363$ & $951$ \\
Ours & $\mathbf{18.145}$ & $\mathbf{15.227}$ & $\mathbf{6512}$ & $\mathbf{14.019}$ & $\mathbf{2.772}$ & $\mathbf{2945}$ \\
Hard to tell & N/A & N/A & $262$ & N/A & N/A & $177$ \\
\hline
\end{tabular}
}
\caption{\textbf{Quantitative Comparison.} We evaluate on our $8,300$ triplets test set and DressCode triplets test set. GP-VTON~\cite{xie2023gp} is trained on layflat garments, thus we report only on DressCode test set. The metrics are FID, KID, and user study (US). All baselines are run twice sequentially, first for tops then for bottoms try-on (See Section~\ref{subsec:mxm_comparison}).}
\label{table:quantitative}
\vspace{-3mm}
\end{table}
\label{subsec:mxm_comparison}
\mypar{Comparison of VTO.} We compared with two representative  state of the art methods: TryOnDiffusion~\cite{Zhu_2023_CVPR_tryondiffusion}, and  GP-VTON~\cite{xie2023gp}. Other methods don't provide code at the time of submission. 
Our  $8,300$ triplets test set was used to compare to TryonDiffusion, and DressCode triplets unpaired test set was used to compare to both GP-VTON and TryonDiffusion.  As TryOnDiffusion was trained only on tops, and person images,  we retrained it on our dataset for upper-body, lower-body, and full-body garments separately. For GP-VTON, we used officially released checkpoints trained on DressCode. Then we ran inference sequentially first to produce top VTO, and then bottom VTO.  Figure~\ref{fig:compare_tryondiffusion} shows that {\mtd} outperforms baselines in aspects such as garment interactions, warping, and detail preservation. Table~\ref{table:quantitative} shows that our method outperforms baselines for FID~\cite{heusel2017gans}, KID~\cite{binkowski2018demystifying} (scaled by $1000$ following~\cite{karras2020training}) and user study (US).  In the user study, 16 non-experts were asked to either select the best result or opt for ``hard to tell."  The findings indicate that users generally prefer {\mtd} over other methods. We  provide  results for single garment try-on, and other comparisons in supplementary.

\mypar{Comparison of Editing.}  We evaluate our approach by comparing with several text-guided image editing methods. Inpainting mask free:  Prompt-to-Prompt (P2P)~\cite{hertz2022prompt} + Null inversion~\cite{mokady2023null} (P2P + NI) and InstructPix2Pix (IP2P)~\cite{brooks2023instructpix2pix} using a target text prompt and an input image that we wish to perform editing on. With inpainting mask:  Imagen editor~\cite{wang2023imagen}, DiffEdit~\cite{couairon2022diffedit} and SDXL inpainting~\cite{podell2023sdxl}. Figure~\ref{fig:edit} demonstrates that our method can interpret garment layout concepts more effectively, allowing for more precise edits of the targeted part without affecting other areas. We provide quantitative comparison and additional details about specific prompts and input masks in supplementary.

\mypar{Finetuning Comparison.} We  compare  to three baselines: non-finetuned model, finetuning the full model and finetuning the person encoder. For the latter two baselines, we have incorporated the class-specific prior preservation loss, as utilized in DreamBooth \cite{ruiz2022dreambooth}, to prevent overfitting to the clothing worn by the target person. For our approach, we don't apply such regularization technique as we found our method does not suffer from  overfitting. Figure~\ref{fig:finetuning} showcases that our method successfully retains characteristics of the human models (\eg, body shape) without compromising the details of the garments.

\begin{figure}[t]
\begin{center}
   \includegraphics[width=1.0\linewidth]{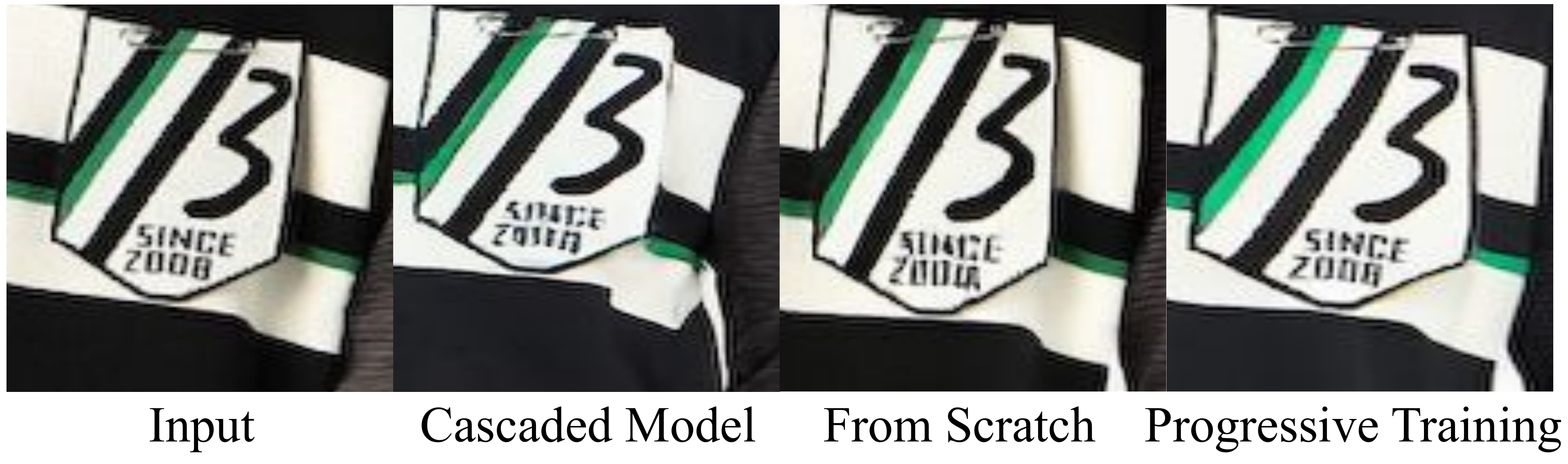}
\end{center}
\vspace{-5mm}
\caption{\textbf{Ablation Comparison}. We provide qualitative zoom-in visualization to compare our progressive training with cascaded models and the model trained from scratch. Our approach can generate better garment details, \ie, more accurate texts. See supplementary for full images.}
\label{fig:ablate_one_stage}
\vspace{-3mm}
\end{figure}
\mypar{Ablation for Single Stage Model vs.\ Cascaded.}  Our method generates  $1024\mathord\times\mathord512$ try-on images in a single stage. For the cascaded variant, we trained a $512\mathord\times\mathord256$ base diffusion model, followed by a $512\mathord\times\mathord256 \rightarrow 1024\mathord\times\mathord512$ SR diffusion model. Both models share the same architecture as our single-stage model, with the distinction that the SR model concatenates the low-resolution image to the noisy image. Figure~\ref{fig:ablate_one_stage} illustrates that our single stage model excels at maintaining complex garment details like tiny texts or logos.

\mypar{Ablation for Progressive Training vs.\ Training from Scratch.} We train an identical model from scratch on $1024\mathord\times\mathord512$ data, without leveraging any model pretrained in lower resolutions. Figure~\ref{fig:ablate_one_stage} highlights that our progressively trained model more effectively manages garment warping under significant pose variations, whereas the ablated version struggles with learning implicit garment warping through cross-attention.
\begin{figure*}[ht]
\begin{center}
   \includegraphics[width=1.0\linewidth]{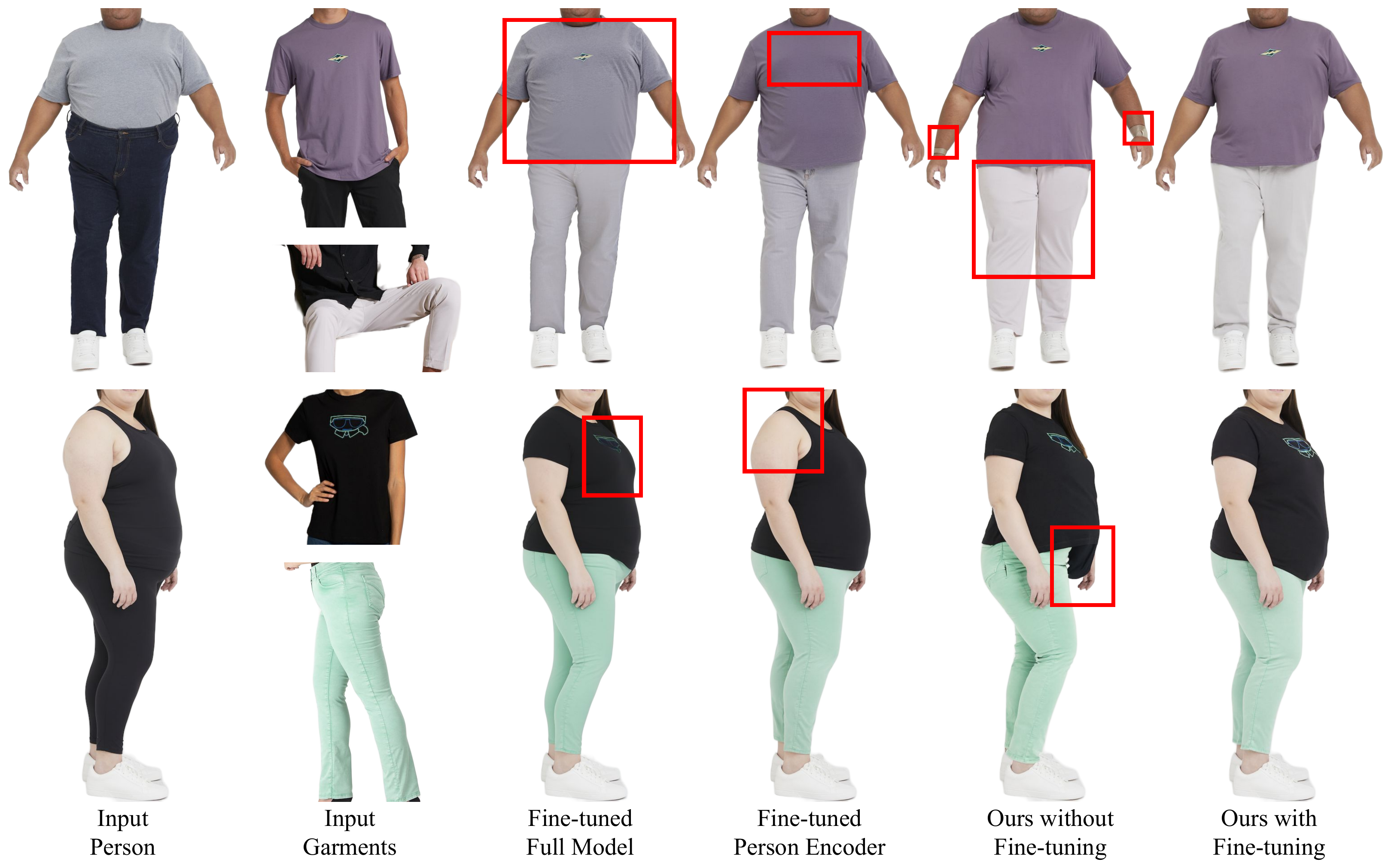}
\end{center}
\vspace{-5mm}
\caption{\textbf{Qualitative Comparison on Person Fine-tuning Strategy.} We  provide a comparison with various types of fine-tuning strategies. Our method shows a better person identity preservation than fine-tuning the whole model or person encoder only. Red boxes highlight example errors, \eg, sleeves too short, and extra fabric. }
\label{fig:finetuning}
\vspace{-3mm}
\end{figure*}

\begin{figure}[t]
\begin{center}
   \includegraphics[width=1\linewidth]{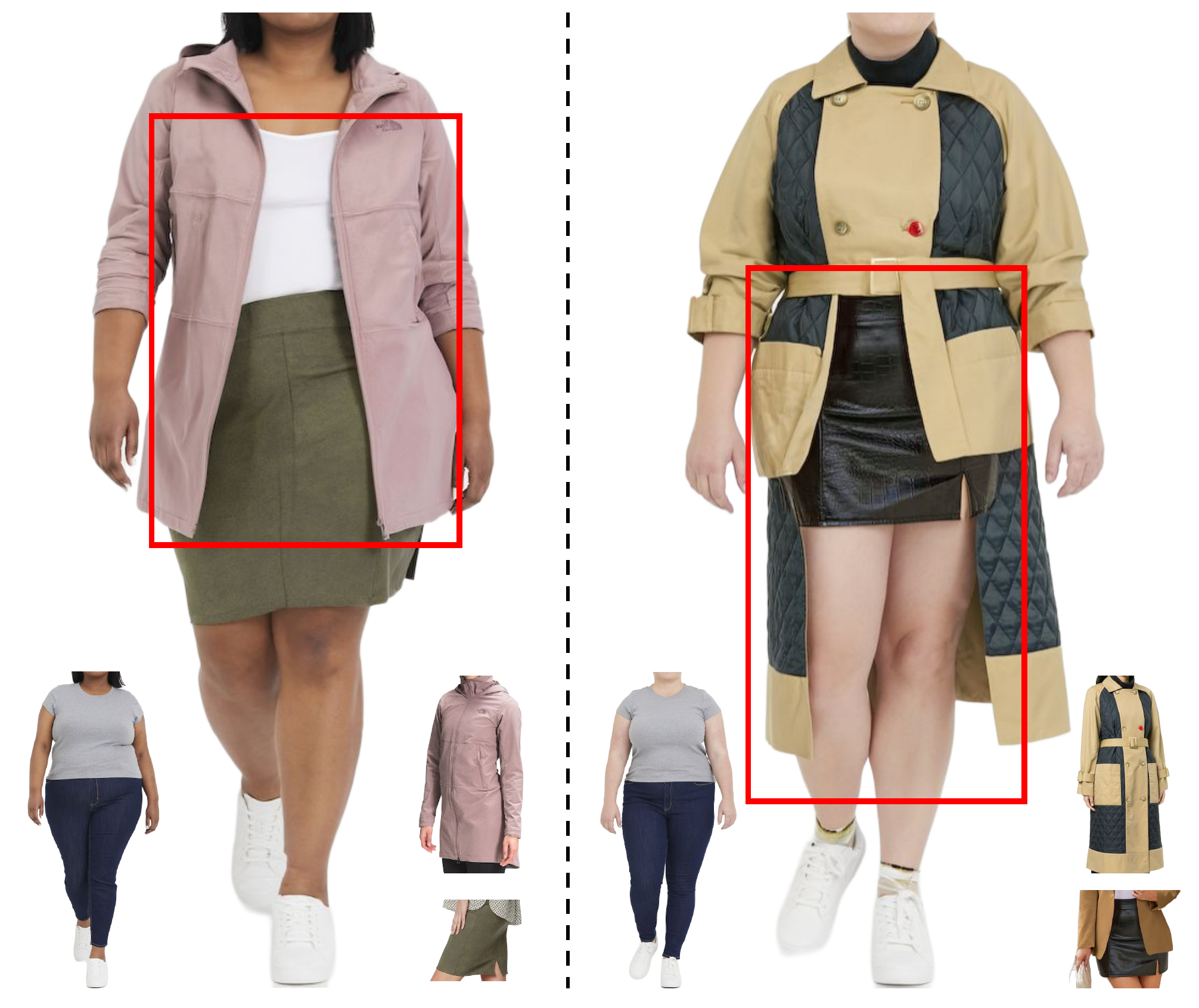}
\end{center}
\vspace{-5mm}
\caption{\textbf{Failure Cases.} Our model could generate random clothing given layout information. As shown in the left example, given ``outer top open'', the model generates a random inner top. In addition, the model could lead to failures when dealing with rare garment combinations. For example, given a long coat and skirt combination, it creates a half open coat, shown in the right image.}
\label{fig:failure_cases}
\vspace{-5mm}
\end{figure}

\mypar{Limitations.} 
Firstly, our approach isn't designed for layout editing tasks, such as ``Open the outer top.'' As demonstrated in Figure~\ref{fig:failure_cases} (left), a random shirt is generated by the model, as no specific information is provided from inputs about what should be inpainted in the open area.
Secondly, our method struggles with uncommon garment combinations found in the real world, like a long coat paired with skirts. As shown in the right example of Figure~\ref{fig:failure_cases}, the model tends to split the long coat in an attempt to show the skirts, because it learned from examples where both garments are typically visible during training.  Thirdly, our model faces challenges when dealing with upper-body clothing from different images, \eg pairing a shirt from one photo with an outer coat from another. This issue mainly stems from the difficulty in finding training pairs where one image clearly shows a shirt without any cover, while another displays the same shirt under an outer layer. As a result, the model struggles to accurately remove the shirt when it's covered by an outer layer during testing.
Finally, note that our method visualizes how an item might look on a person, accounting for their body shape, but it doesn't yet include size information nor solves for exact fit. 
\section{Conclusion}
\label{sec:conclusion}

We present a method that can synthesize multi-garment try-on results given an image of person and images of upper-body, lower-body and full-body garments. Our  novel architecture VTO-UDiT as well as  progressive training strategy, enabled better than state-of-the-art results, particularly in preserving fine garment details and person identity. Furthermore, our method allows for  explicit control of  garment layout via conditioning the model with garment attributes obtained from a finetuned vision-language model. 

{
    \small
    \bibliographystyle{ieeenat_fullname}
    \bibliography{main}
}

\clearpage
\maketitlesupplementary

\section{Implementation Details}
\label{sec:implementation_details}

\subsection{Training and inference}
{\mtd} is trained in two stages.
For the first stage, the model is trained on $512\mathord\times\mathord256$ images for $600K$ iterations. In the second stage, the model is initialized from the pretrained checkpoint of the first stage and trained on $1024\mathord\times\mathord512$ images for an additional $200K$ iterations. For both training stages, the batch size is set to $1024$, and the learning rate linearly increases from $0$ to $10^{-4}$ in the first $10K$ steps and is kept unchanged afterwards. We parameterize the model output in $v$-space following~\cite{salimans2022progressive} while the $L2$ loss is computed in $\epsilon$-space. All conditional inputs are set to 0 in $10\%$ of the training time for classifier-free guidance (CFG) \cite{ho2022classifier}.
Test results are generated by sampling {\mtd} for $256$ steps using ancestral sampler~\cite{ho2020denoising}.

\subsection{Garment attributes}

We summarize as follows the full set of attributes used as layout conditioning input $y_{\text{gl}}$.
\begin{enumerate}
  \item What is the type of the sleeve?
  \begin{enumerate}
    \item Not applicable
    \item Sleeveless
    \item Short sleeve
    \item Middle sleeve
    \item Long sleeve
  \end{enumerate}
  \item Is the sleeve rolled up?
  \begin{enumerate}
    \item Not applicable
    \item Sleeve type is not long
    \item Yes
    \item No
  \end{enumerate}
  \item Is the top garment tucked in?
  \begin{enumerate}
    \item Not applicable
    \item Not wearing top garment
    \item Can not determine
    \item Yes
    \item No
  \end{enumerate}
  \item Is the person wearing outer top?
  \begin{enumerate}
    \item Not applicable
    \item Yes
    \item No
  \end{enumerate}
  \item Is the outer top closed (\eg zipper up or button on)?
  \begin{enumerate}
    \item Not applicable
    \item Not wearing outer top
    \item Can not determine
    \item Yes
    \item No
  \end{enumerate}
\end{enumerate}

We selected $1,500$ images and asked human labelers to answer all questions for each image.
After that, we converted question-answer pairs into a formatted text, where different question-answer pairs are separated by semicolon while the question and answer within each pair are separated by 
colon. The resulting $1,500$ image-caption samples were used to finetune PaLI-3~\cite{chen2023pali} model. Finally, we ran inference of the finetuned model on our train and test data, and converted the formatted text back into class labels.

\section{Results}

In this section, we provide additional qualitative and quantitative results.

\subsection{Comparison of VTO}
\begin{figure*}[htb]
\begin{center}
   \includegraphics[width=0.9\linewidth]{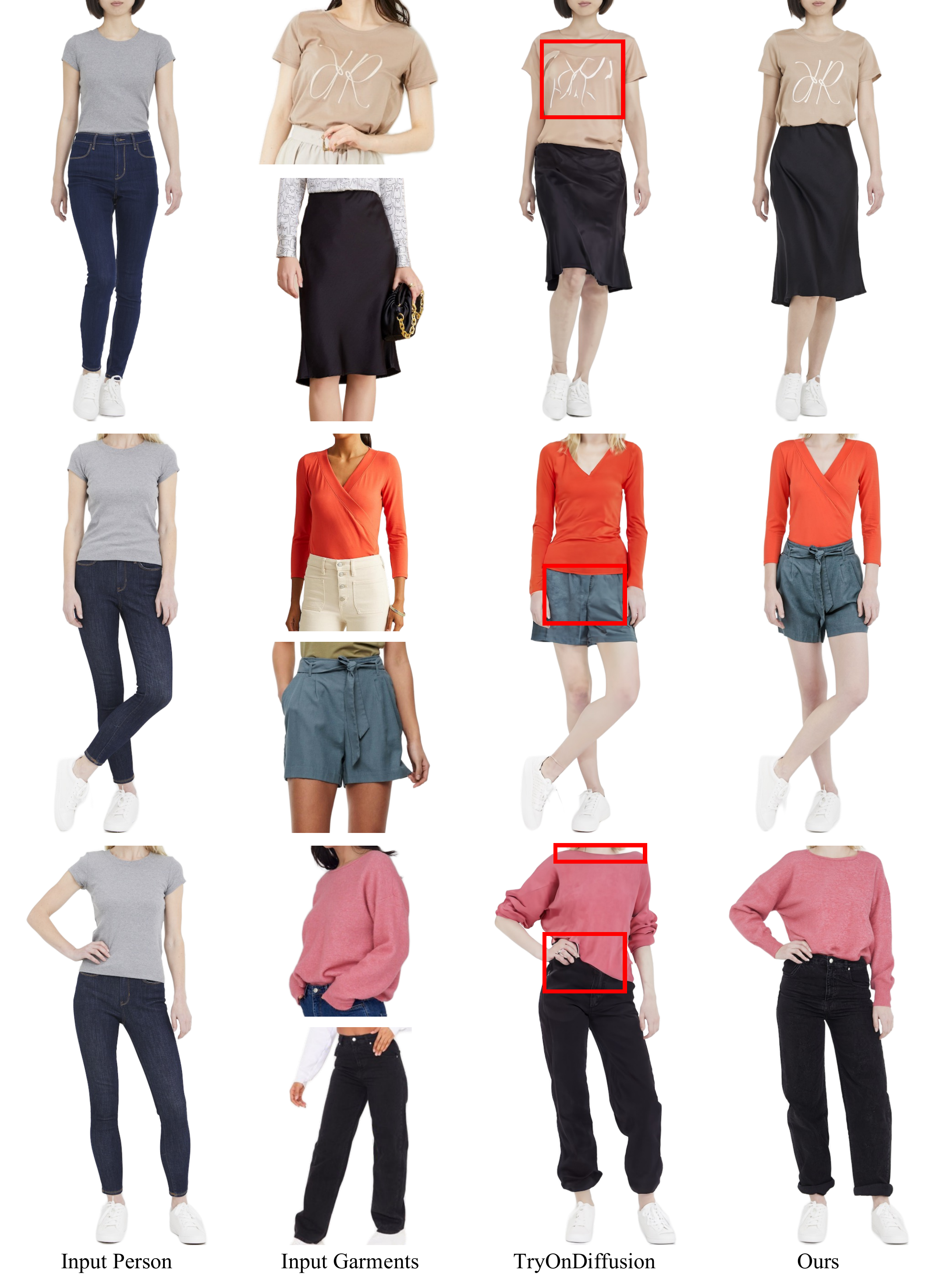}
\end{center}
\vspace{-8mm}
\caption{\textbf{Qualitative comparison against TryOnDiffusion~\cite{Zhu_2023_CVPR_tryondiffusion} on our $\mathbf{8,300}$ triplets test set part one.} Our method can generate better garment details and layouts. Red boxes highlight errors of TryOnDiffusion. Please zoom in to see details.}
\label{fig:suppl_tryondiffusion_8300_women1}
\end{figure*}

\begin{figure*}[htb]
\begin{center}
   \includegraphics[width=0.9\linewidth]{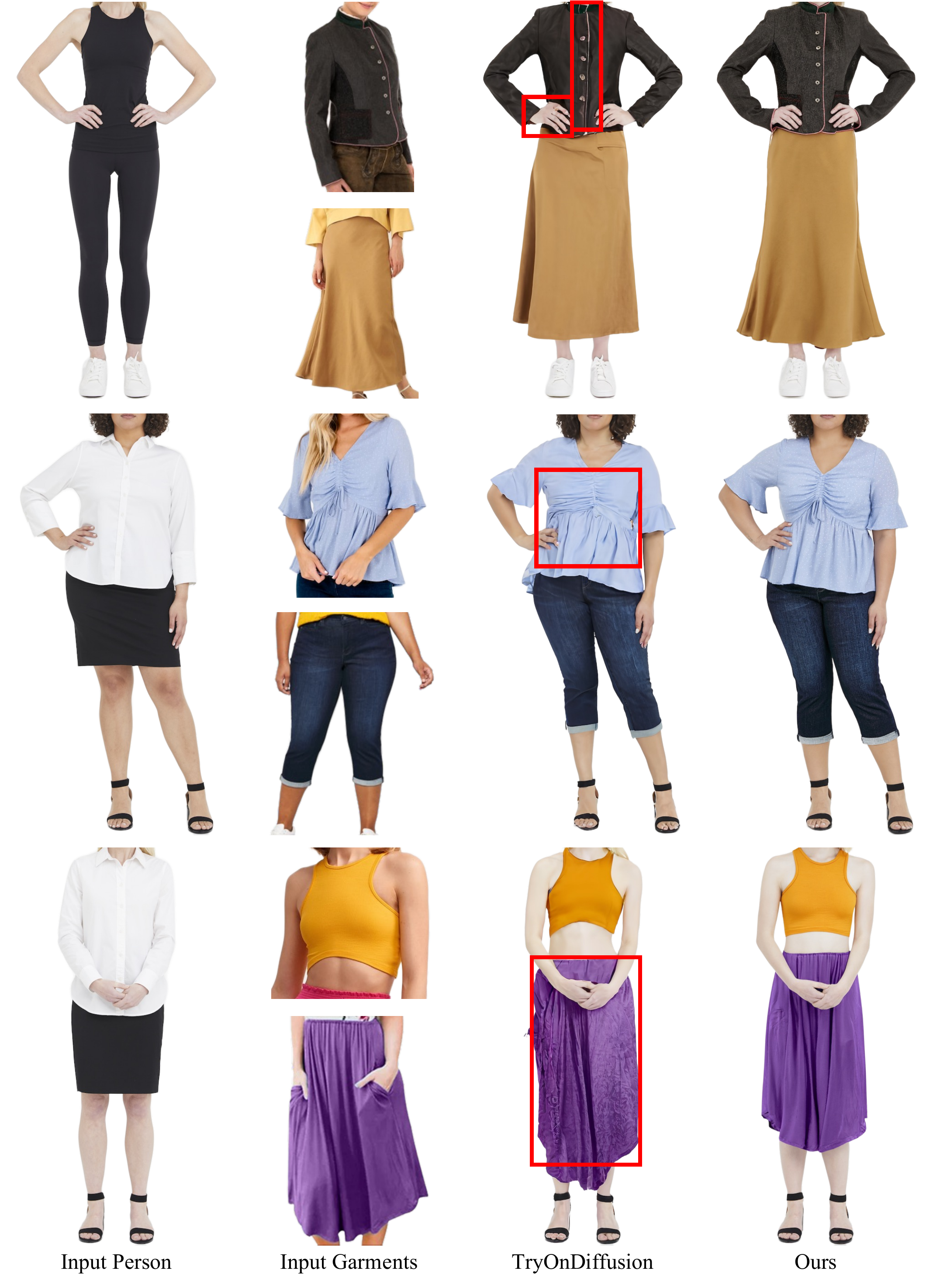}
\end{center}
\vspace{-8mm}
\caption{\textbf{Qualitative comparison against TryOnDiffusion~\cite{Zhu_2023_CVPR_tryondiffusion} on our $\mathbf{8,300}$ triplets test set part two.} Our method can generate better garment details and layouts. Red boxes highlight errors of TryOnDiffusion. Please zoom in to see details.}
\label{fig:suppl_tryondiffusion_8300_women2}
\end{figure*}

\begin{figure*}[htb]
\begin{center}
   \includegraphics[width=0.9\linewidth]{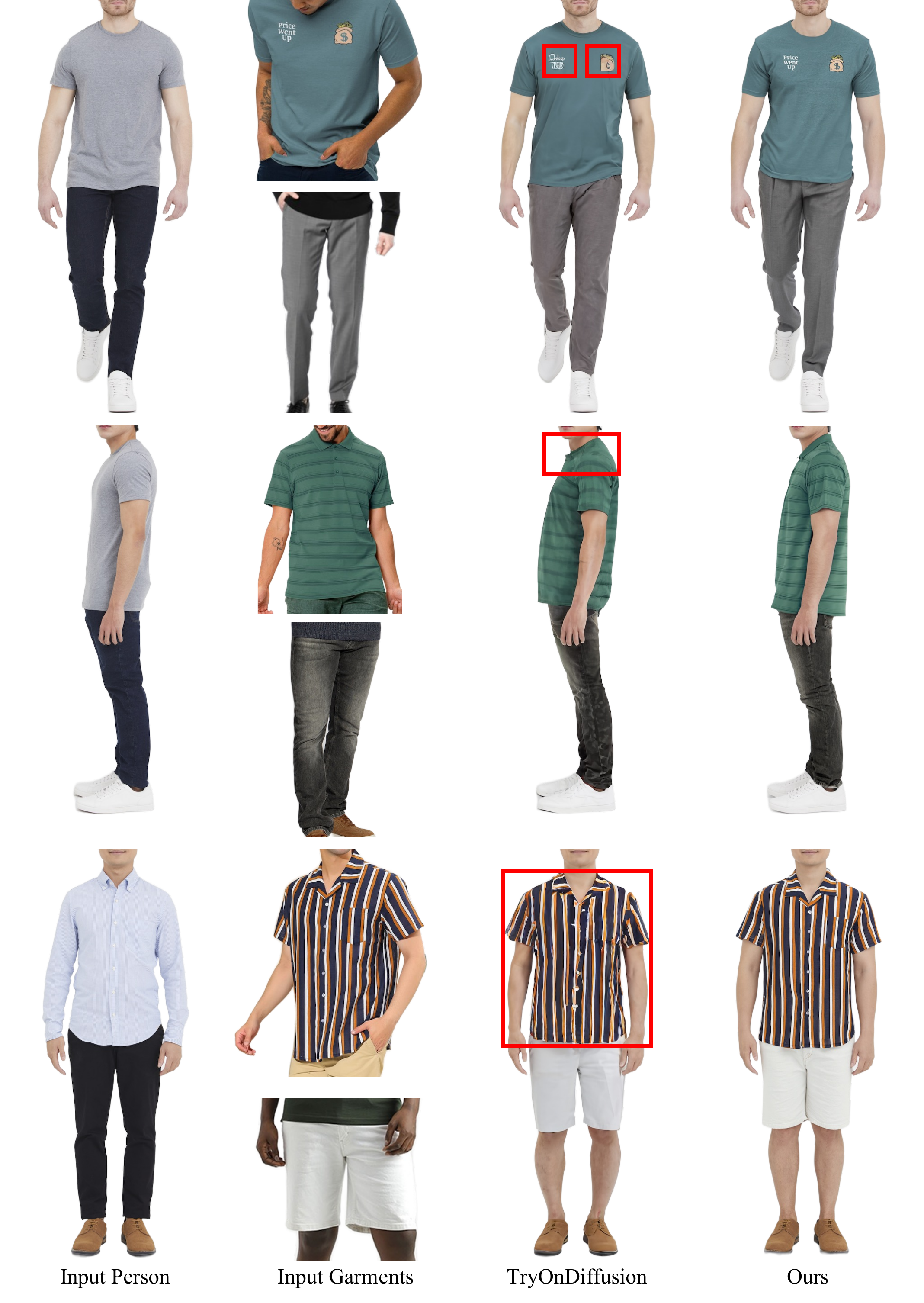}
\end{center}
\vspace{-8mm}
\caption{\textbf{Qualitative comparison against TryOnDiffusion~\cite{Zhu_2023_CVPR_tryondiffusion} on our $\mathbf{8,300}$ triplets test set part three.} Our method can generate better garment details and layouts. Red boxes highlight errors of TryOnDiffusion. Please zoom in to see details.}
\label{fig:suppl_tryondiffusion_8300_men1}
\end{figure*}

\begin{figure*}[htb]
\begin{center}
   \includegraphics[width=0.9\linewidth]{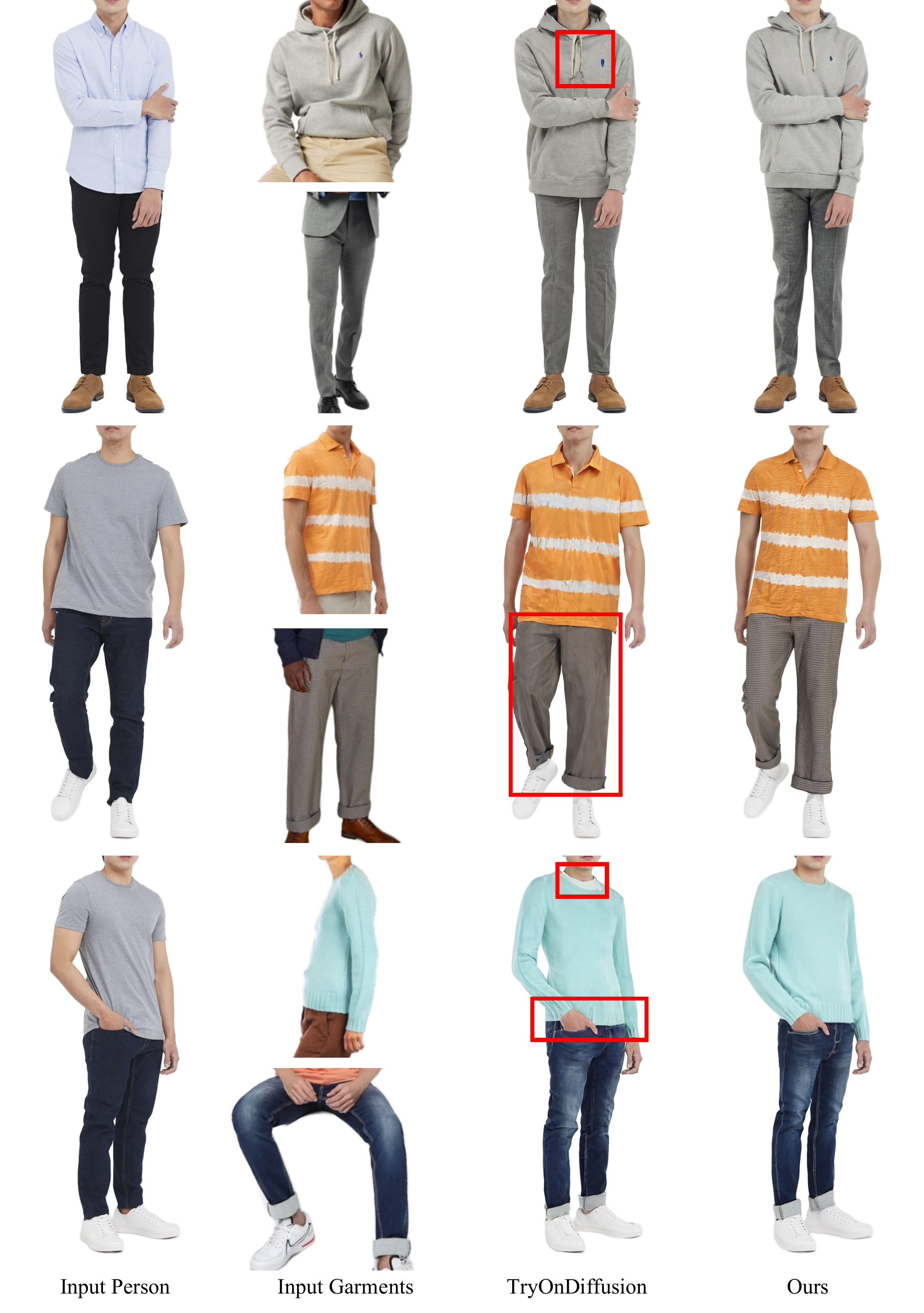}
\end{center}
\vspace{-8mm}
\caption{\textbf{Qualitative comparison against TryOnDiffusion~\cite{Zhu_2023_CVPR_tryondiffusion} on our $\mathbf{8,300}$ triplets test set part four.} Our method can generate better garment details and layouts. Red boxes highlight errors of TryOnDiffusion. Please zoom in to see details.}
\label{fig:suppl_tryondiffusion_8300_men2}
\end{figure*}

\begin{figure*}[htb]
\begin{center}
   \includegraphics[width=1\linewidth]{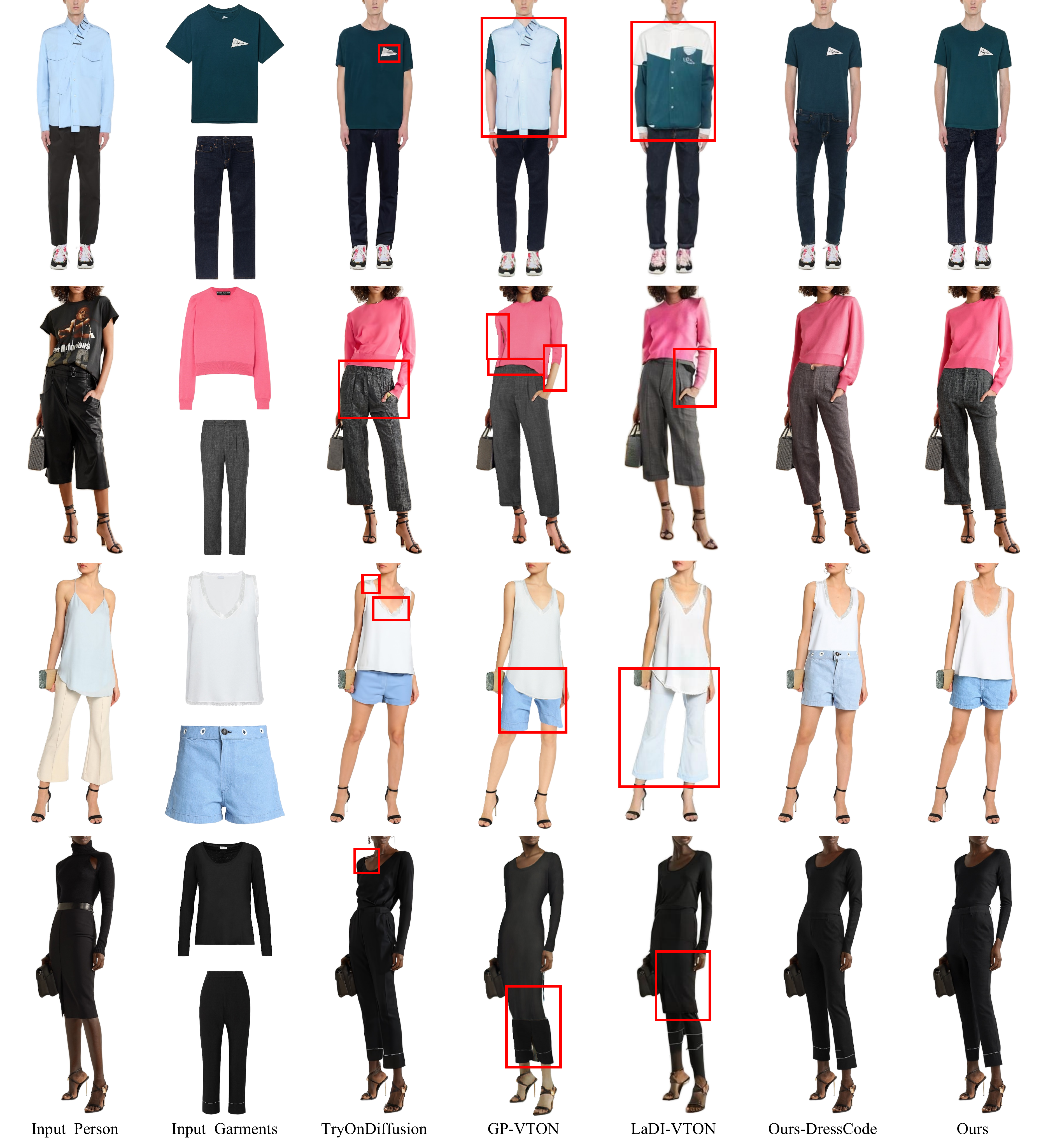}
\end{center}
\vspace{-8mm}
\caption{\textbf{Qualitative comparison against GP-VTON~\cite{xie2023gp}, LaDI-VTON~\cite{morelli2023ladi} and TryOnDiffusion~\cite{Zhu_2023_CVPR_tryondiffusion} on DressCode\cite{morelli2022dress} triplets test set part one.} Ours-DressCode represents our method trained only on DressCode dataset. Red boxes highlight errors of baselines. Please zoom in to see details.}
\label{fig:suppl_dresscode_1}
\end{figure*}

\begin{figure*}[htb]
\begin{center}
   \includegraphics[width=0.95\linewidth]{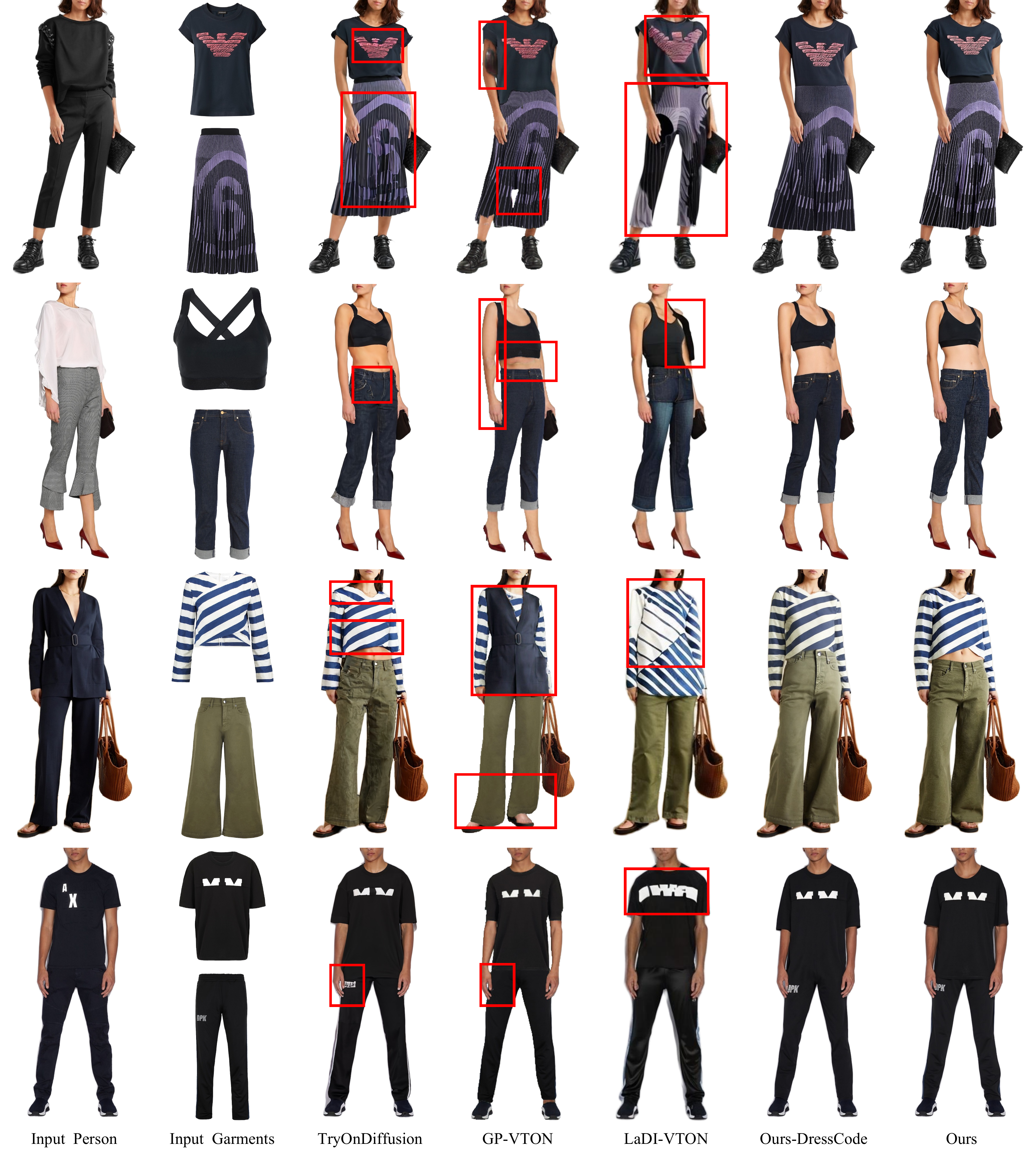}
\end{center}
\vspace{-8mm}
\caption{\textbf{Qualitative comparison against GP-VTON~\cite{xie2023gp}, LaDI-VTON~\cite{morelli2023ladi} and TryOnDiffusion~\cite{Zhu_2023_CVPR_tryondiffusion} on DressCode\cite{morelli2022dress} triplets test set part two.} Ours-DressCode represents our method trained only on DressCode dataset. Red boxes highlight errors of baselines. Please zoom in to see details.}
\label{fig:suppl_dresscode_2}
\end{figure*}

\begin{table}[t]
\centering
\scalebox{0.7}{
\begin{tabular}{ |c|cc|  }
\hline
Methods & FID $\downarrow$ & KID $\downarrow$ \\ 
\hline 
GP-VTON~\cite{xie2023gp} & $38.392$ & $33.909$ \\
LaDI-VTON~\cite{morelli2023ladi} & $19.346$ & $9.305$ \\
Ours-DressCode & $\mathbf{18.725}$ & $\mathbf{8.250}$  \\
\hline
\end{tabular}
}
\caption{Our method trained solely on DressCode vs GP-VTON and LaDI-VTON official checkpoints. We report FID and KID on DressCode triplets test set.}
\label{table:dresscode_solely}
\vspace{-3mm}
\end{table}
In Figure~\ref{fig:suppl_tryondiffusion_8300_women1},~\ref{fig:suppl_tryondiffusion_8300_women2},~\ref{fig:suppl_tryondiffusion_8300_men1} and ~\ref{fig:suppl_tryondiffusion_8300_men2}, we showcase additional qualitative results from our $8,300$ triplets test set, comparing them against those generated by TryOnDiffusion~\cite{Zhu_2023_CVPR_tryondiffusion}, where both methods are trained on our ``garment paired'' and ``layflat paired'' dataset. These results highlight our method's superior ability to retain garment details and layout. We also compare to ``layflat-VTO'' methods GP-VTON~\cite{xie2023gp} and LaDI-VTON~\cite{morelli2023ladi} on DressCode~\cite{morelli2022dress} triplets test dataset. To ensure a fair comparison, we trained our method exclusively on the DressCode dataset. The FID and KID metrics for the DressCode triplets test set, presented in Table~\ref{table:dresscode_solely}, demonstrate that our method surpasses GP-VTON and LaDI-VTON in both metrics, even when \textbf{trained solely} on the DressCode dataset. Further qualitative comparisons on the DressCode triplets test set against all baselines are provided in Figure~\ref{fig:suppl_dresscode_1} and ~\ref{fig:suppl_dresscode_2}.

\subsection{Comparison of Editing}
\begin{figure*}[htb]
\begin{center}
   \includegraphics[width=1\linewidth]{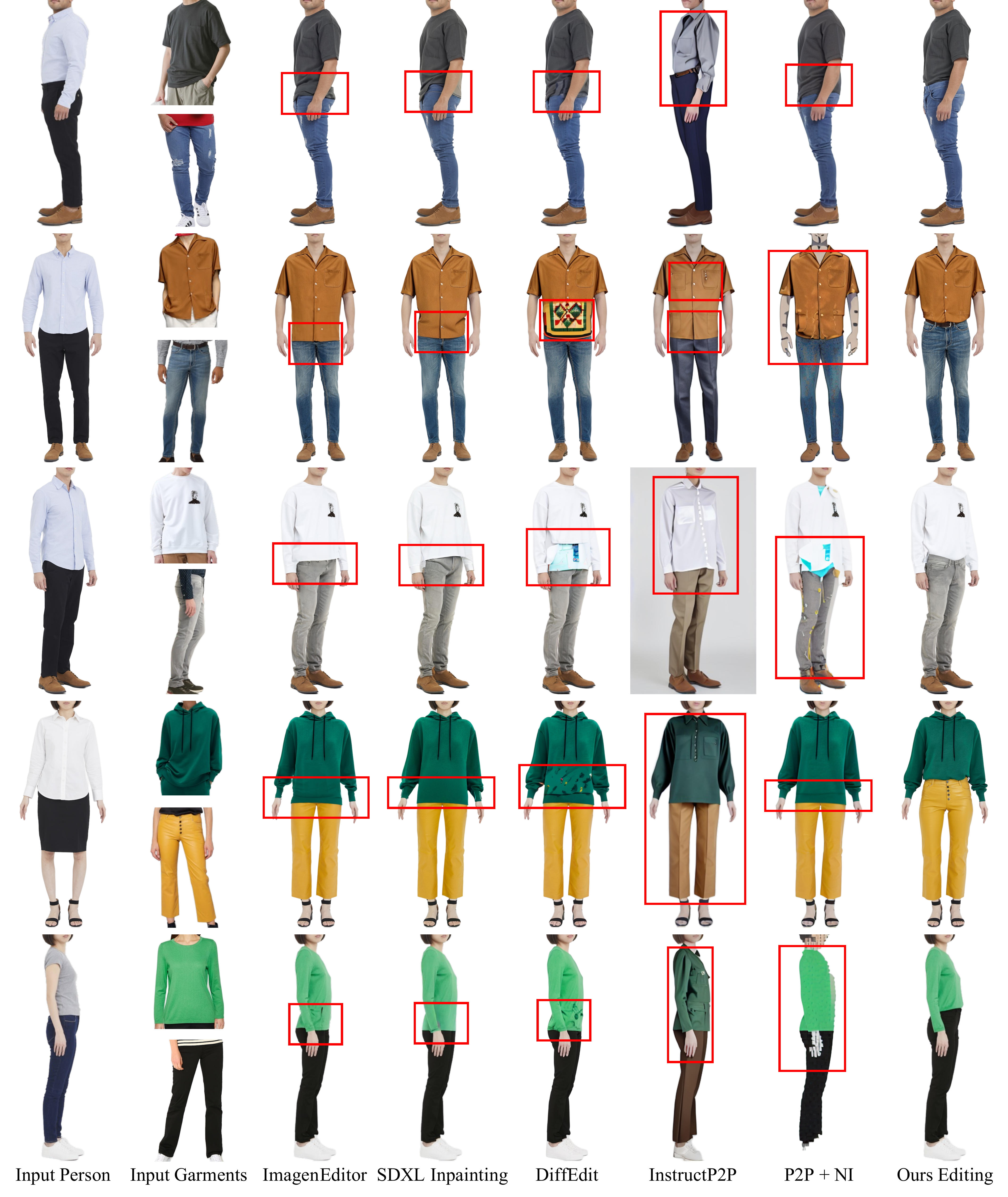}
\end{center}
\vspace{-5mm}
\caption{\textbf{Qualitative comparison for editing instruction: ``tuck in the shirt``.} Please zoom in to see how our method can perform the desired editing while preserving garment details. Red boxes highlight errors of baselines.}
\label{fig:suppl_make_tucked_in}
\end{figure*}

\begin{figure*}[htb]
\begin{center}
   \includegraphics[width=1\linewidth]{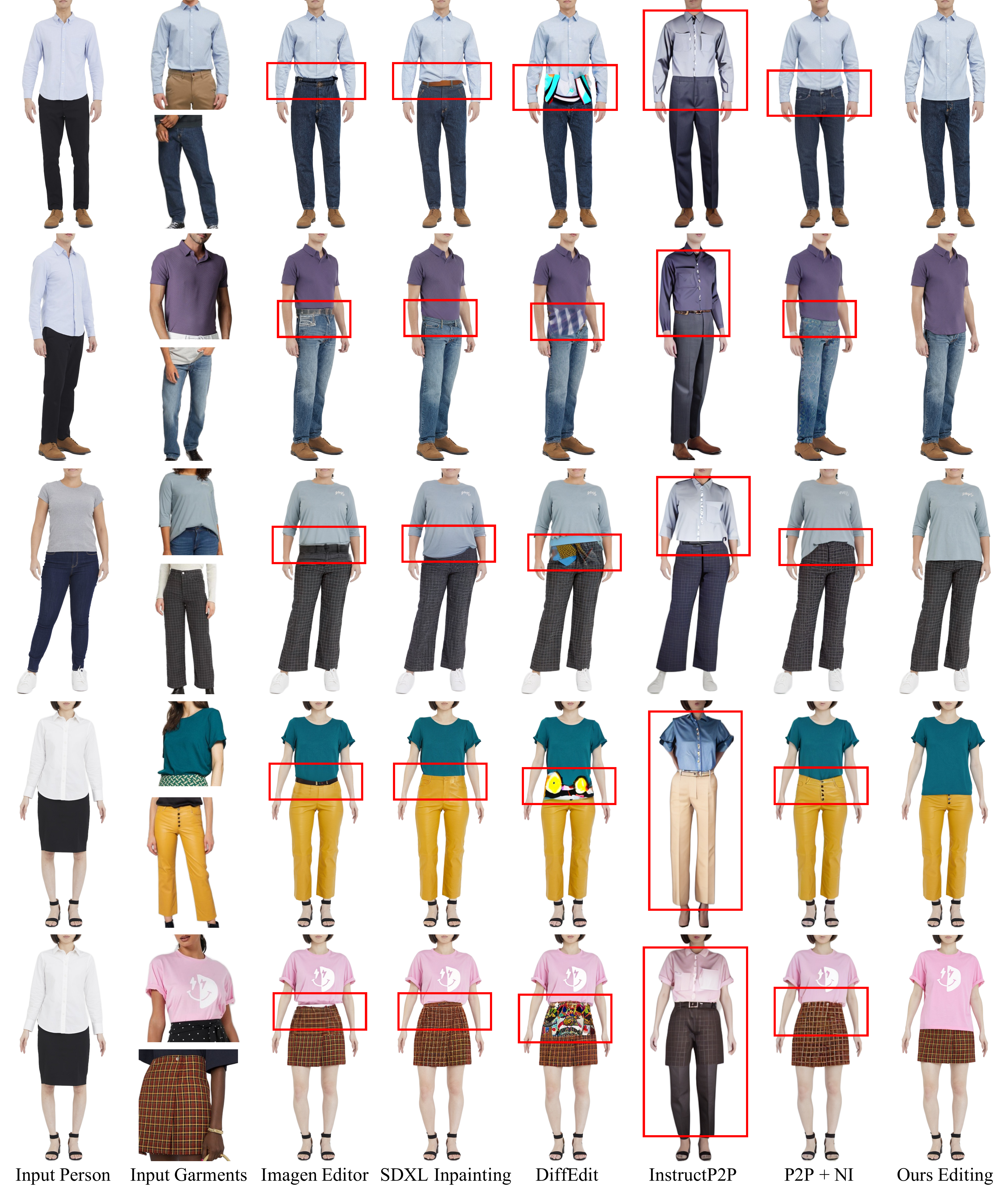}
\end{center}
\vspace{-5mm}
\caption{\textbf{Qualitative comparison for editing instruction: ``tuck out the shirt``.} Please zoom in to see how our method can perform the desired editing while preserving garment details. Red boxes highlight errors of baselines.}
\label{fig:suppl_make_tucked_out}
\end{figure*}

\begin{figure*}[htb]
\begin{center}
   \includegraphics[width=1\linewidth]{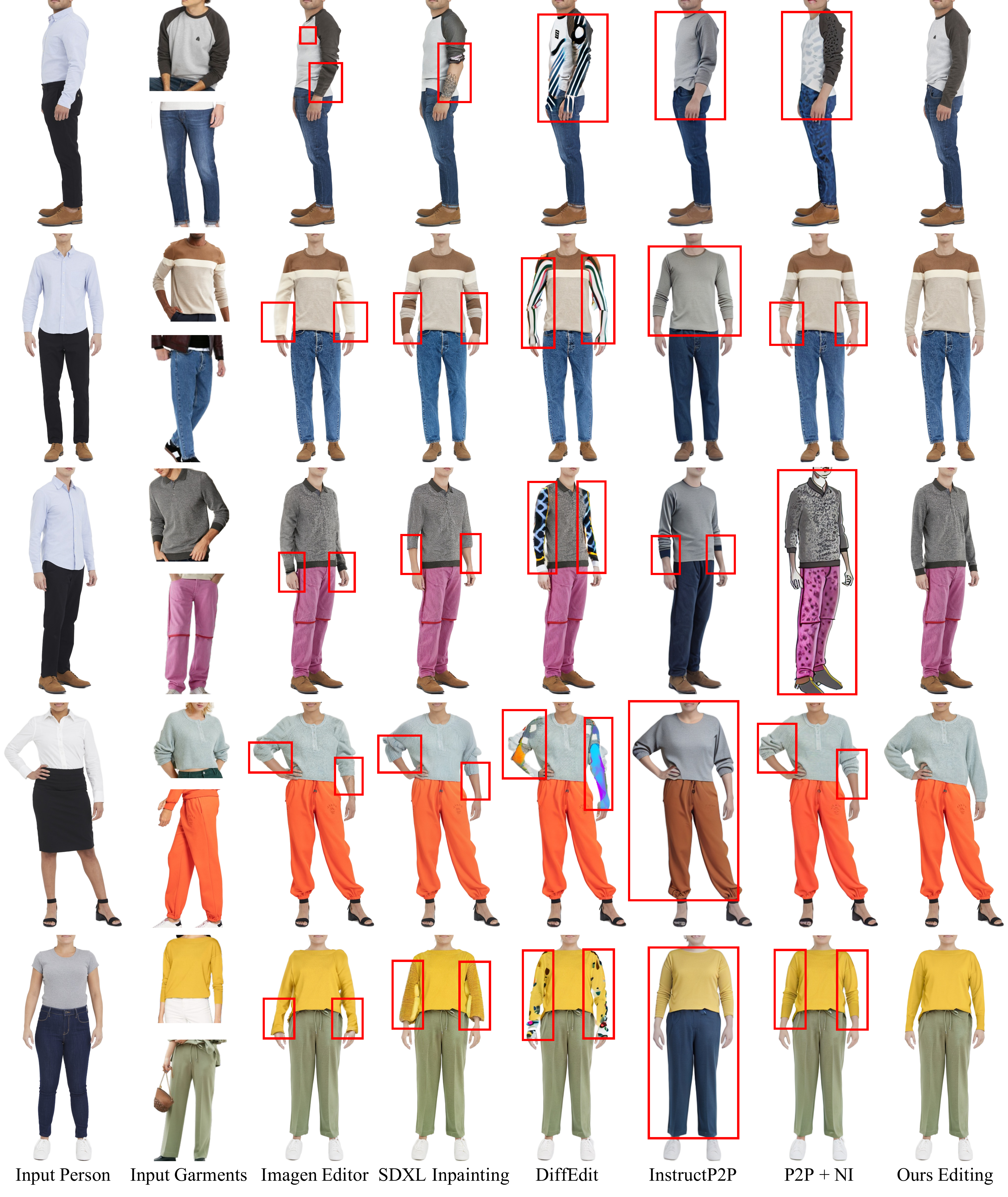}
\end{center}
\vspace{-5mm}
\caption{\textbf{Qualitative comparison for editing instruction: ``roll down the sleeve``.} Please zoom in to see how our method can perform the desired editing while preserving garment details. Red boxes highlight errors of baselines.}
\label{fig:suppl_make_rolled_down}
\end{figure*}

\begin{figure*}[htb]
\begin{center}
   \includegraphics[width=1\linewidth]{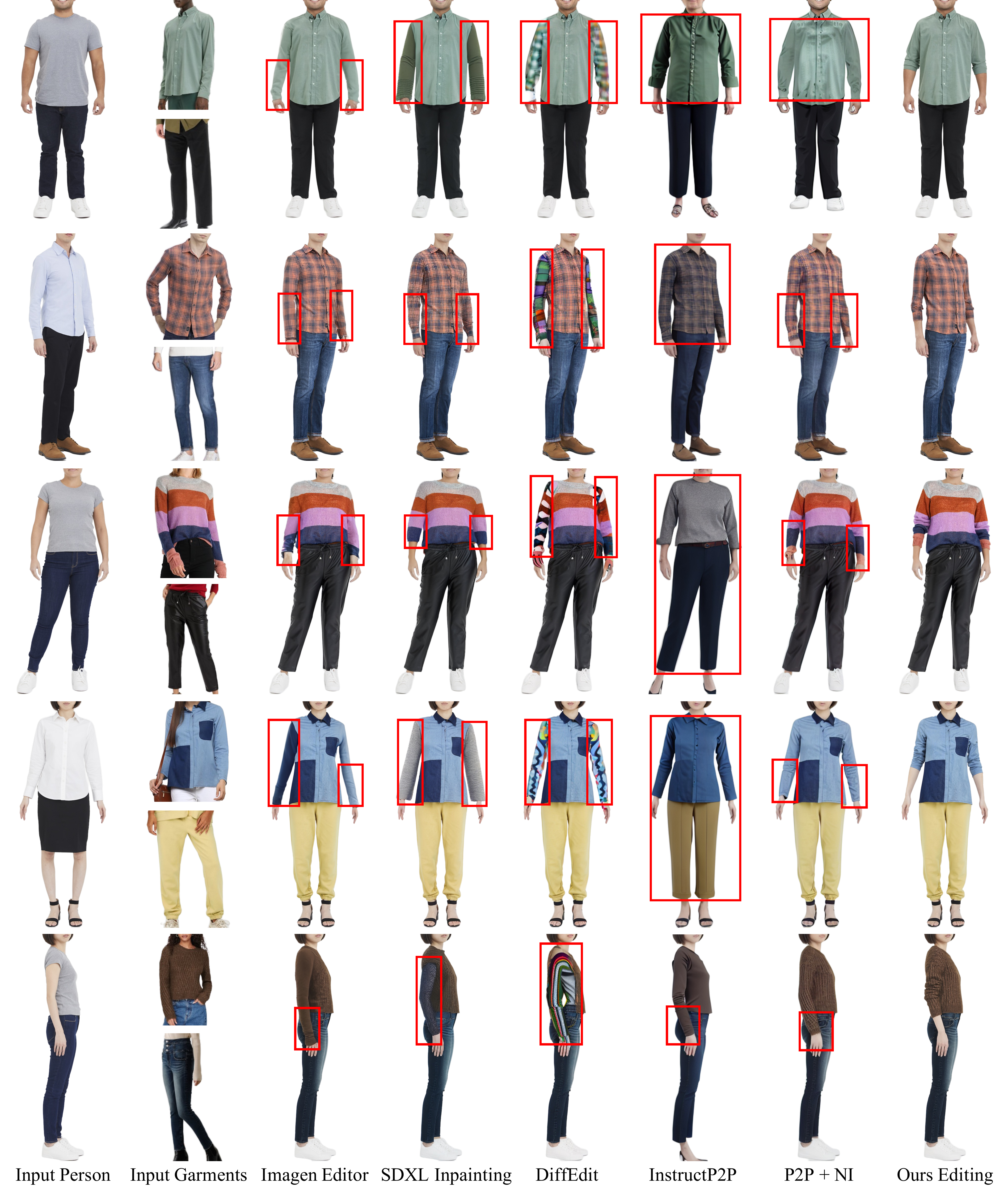}
\end{center}
\vspace{-5mm}
\caption{\textbf{Qualitative comparison for editing instruction: ``roll up the sleeve``.} Please zoom in to see how our method can perform the desired editing while preserving garment details. Red boxes highlight errors of baselines.}
\label{fig:suppl_make_rolled_up}
\end{figure*}

\begin{table}[t]
\centering
\scalebox{0.7}{
\begin{tabular}{ |cc| }
\hline
Methods & US $\uparrow$ \\ 
\hline 
P2P + NI~\cite{mokady2023null} & $0$ \\
IP2P~\cite{brooks2023instructpix2pix} & $1$ \\
Imagen editor~\cite{wang2023imagen} & $10$\\
DiffEdit~\cite{couairon2022diffedit} & $0$ \\
SDXL inpainting~\cite{podell2023sdxl} & $4$ \\
Ours & $\mathbf{169}$ \\
Hard to tell & $16$ \\
\hline
\end{tabular}
}
\caption{\textbf{User Study for try-on editing.} We conducted user study on 200 images. The users are required to select the best method that can successfully perform the editing task while maintaining the property of input person and garments.}
\label{table:edit_qa}
\vspace{-3mm}
\end{table}
We conducted a user study with $200$ images to compare garment layout editing. The results in Table~\ref{table:edit_qa} indicate that our method are preferred by users $84.5\%$ of the time, outperforming the baseline methods. Figure~\ref{fig:suppl_make_tucked_in},~\ref{fig:suppl_make_tucked_out},~\ref{fig:suppl_make_rolled_down} and ~\ref{fig:suppl_make_rolled_up} present qualitative comparisons on different layout editing tasks. These examples demonstrate our method's ability to perform the intended edits accurately while preserving the integrity of other areas in both the person and the garments.

Image editing baselines require different sets of inputs, such as masks. InstructPix2Pix~\cite{brooks2023instructpix2pix} and Prompt-to-Prompt (P2P)~\cite{hertz2022prompt} with null inversion~\cite{mokady2023null} only requires text editing instructions. DiffEdit~\cite{couairon2022diffedit}, Imagen Editor~\cite{wang2023imagen}, and Stable Diffusion XL Inpainting~\cite{podell2023sdxl} require masks for the region of interest. To automatically obtain masks for image editing, we use human pose estimations to mask out belly regions for ``tuck in top garment'' or ``tuck out top garment'' or the arm regions for ``roll up sleeve'' or ``roll down sleeve''. 

\subsection{Finetuning Comparison}
\begin{figure*}[htb]
\begin{center}
   \includegraphics[width=1\linewidth]{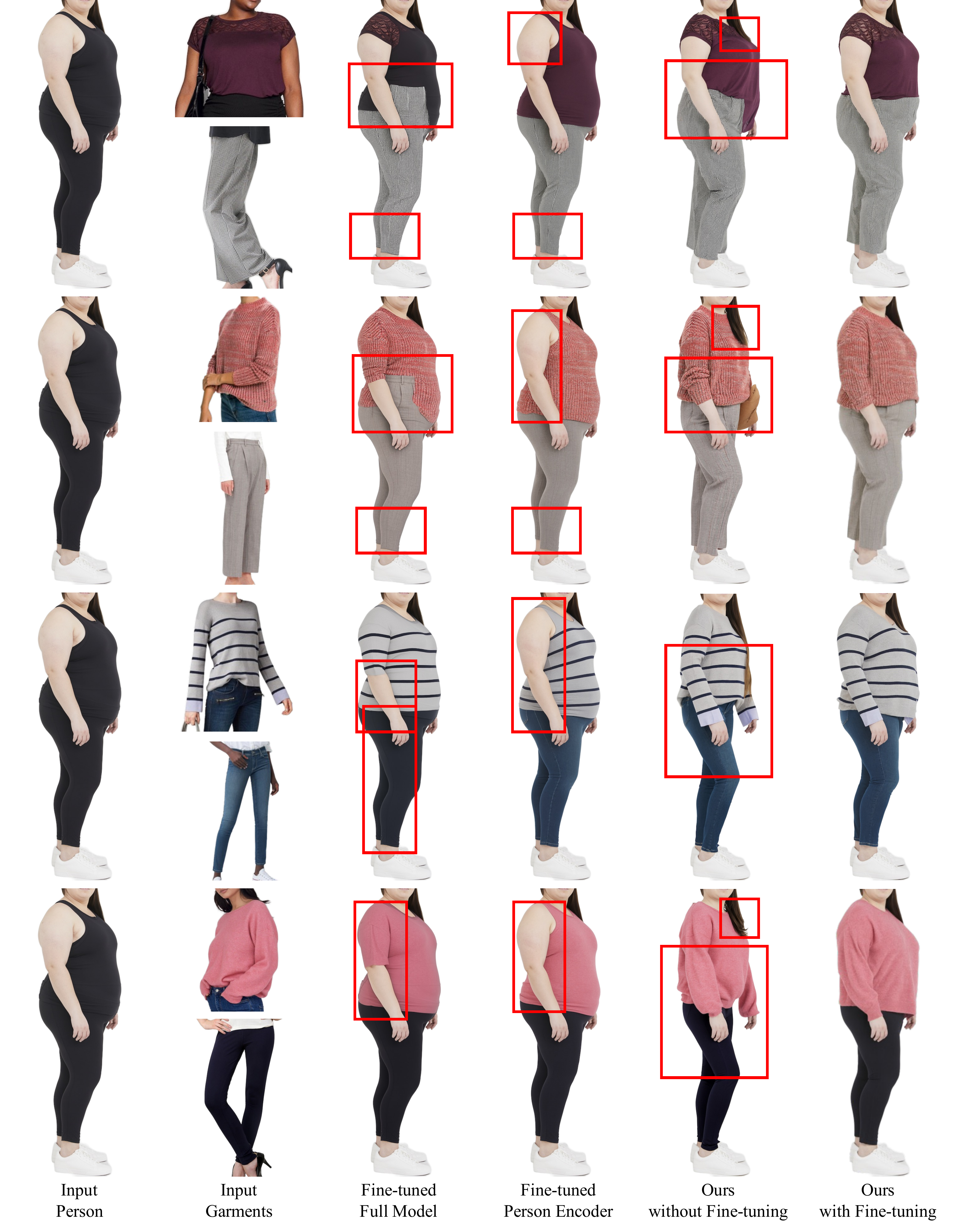}
\end{center}
\vspace{-8mm}
\caption{\textbf{Qualitative comparison for person finetuning of subject $\mathbf{1}$.} Please zoom in to see how our method can preserve both person identity and garment details. Red boxes highlight errors of baselines.}
\label{fig:suppl_finetune_subject1}
\end{figure*}

\begin{figure*}[htb]
\begin{center}
   \includegraphics[width=1\linewidth]{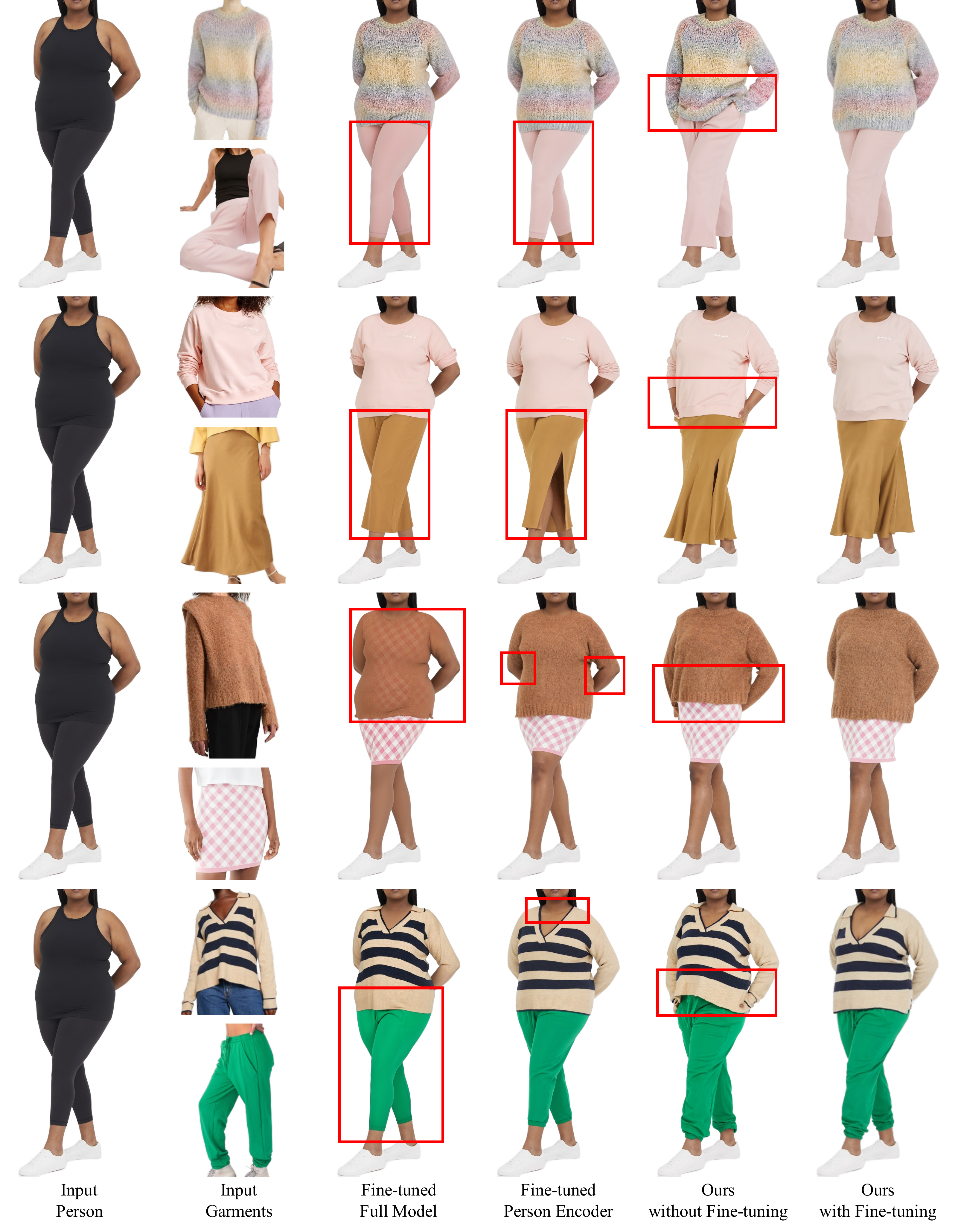}
\end{center}
\vspace{-8mm}
\caption{\textbf{Qualitative comparison for person finetuning of subject $\mathbf{2}$.} Please zoom in to see how our method can preserve both person identity and garment details. Red boxes highlight errors of baselines.}
\label{fig:suppl_finetune_subject2}
\end{figure*}

\begin{figure*}[htb]
\begin{center}
   \includegraphics[width=1\linewidth]{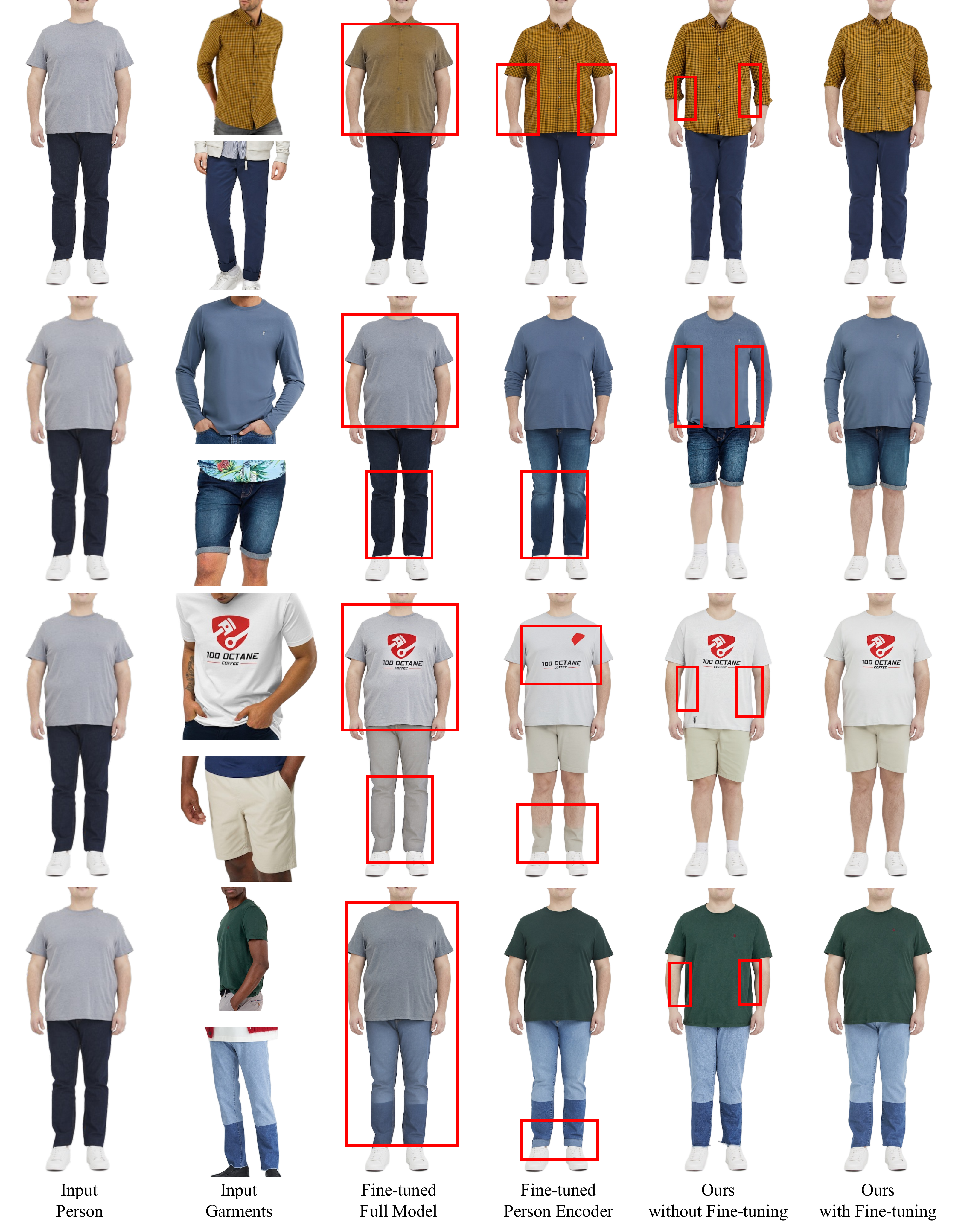}
\end{center}
\vspace{-8mm}
\caption{\textbf{Qualitative comparison for person finetuning of subject $\mathbf{3}$.} Please zoom in to see how our method can preserve both person identity and garment details. Red boxes highlight errors of baselines.}
\label{fig:suppl_finetune_subject3}
\end{figure*}

\begin{figure*}[htb]
\begin{center}
   \includegraphics[width=1\linewidth]{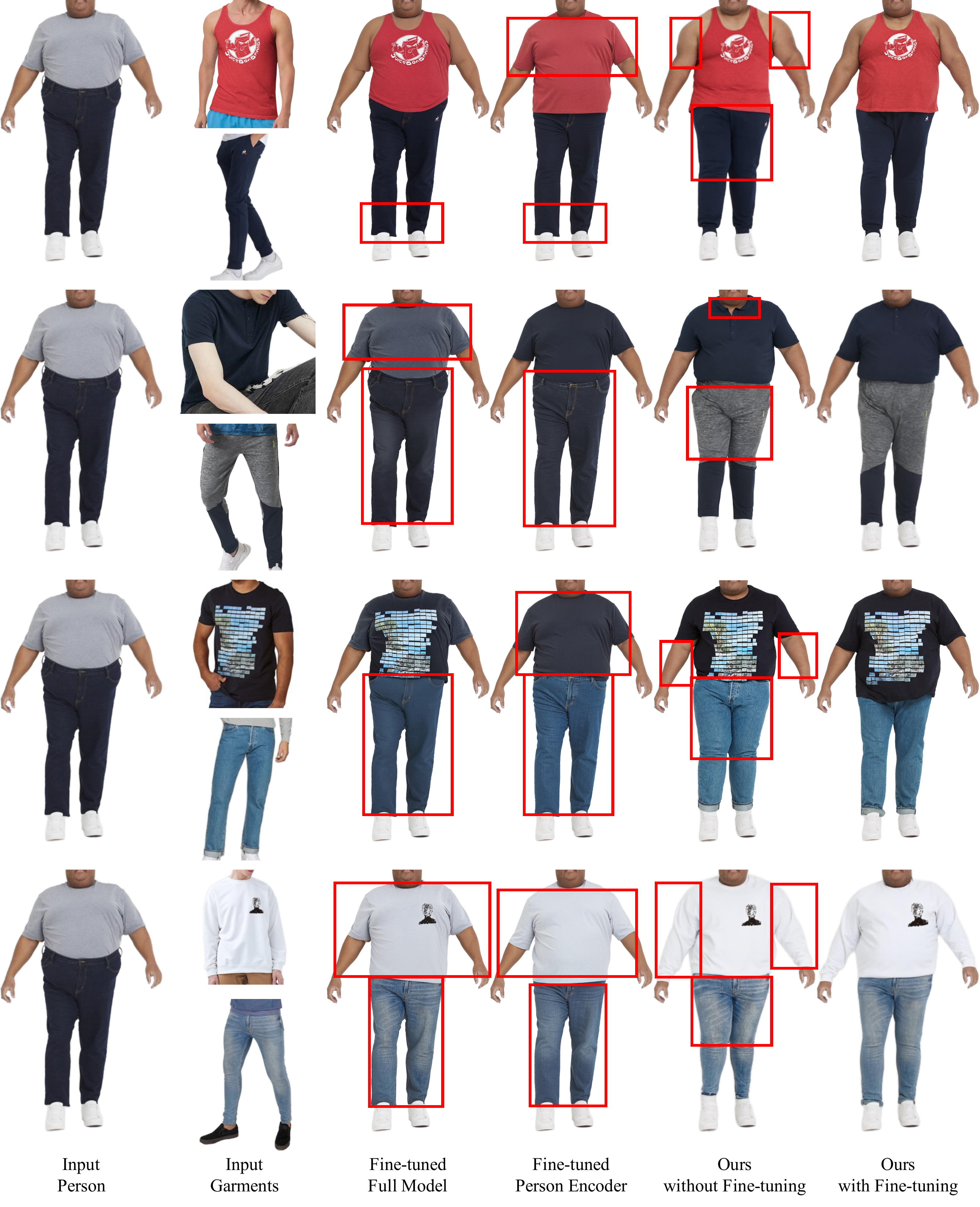}
\end{center}
\vspace{-8mm}
\caption{\textbf{Qualitative comparison for person finetuning of subject $\mathbf{4}$.} Please zoom in to see how our method can preserve both person identity and garment details. Red boxes highlight errors of baselines.}
\label{fig:suppl_finetune_subject4}
\end{figure*}

\begin{table}[t]
\centering
\scalebox{0.7}{
\begin{tabular}{ |cc| }
\hline
Methods & US $\uparrow$ \\ 
\hline 
Finetuned full model & $19$ \\
Finetuned person encoder  & $20$\\
Ours without finetuning & $95$ \\
Ours with finetuning & $\mathbf{265}$ \\
Hard to tell & $1$ \\
\hline
\end{tabular}
}
\caption{\textbf{User Study for person finetuning.} We carried out a user study involving $400$ images across $4$ subjects, where we randomly select $100$ top + bottom input garments for each subject. The participants were asked to choose the method that best maintains the identity of the person (including body pose and shape) as well as the details of input garments.}
\label{table:finetune_qa}
\vspace{-3mm}
\end{table}
We chose $4$ person images with challenging body shapes or poses for our person finetuning comparison. For each person image, we randomly picked $100$ top and bottom garment combinations, then generated try-on results using all baseline methods as well as our own. The user study results, detailed in Table~\ref{table:finetune_qa}, show our finetuning method significantly outperforming the baselines. Additionally, Figure~\ref{fig:suppl_finetune_subject1},~\ref{fig:suppl_finetune_subject2},~\ref{fig:suppl_finetune_subject3} and \ref{fig:suppl_finetune_subject4} showcase qualitative comparison for each subject. Without finetuning, the person's arms, legs, or torso may appear unnaturally slim or wide, and certain challenging poses can not be accurately recovered. However, if we finetune the entire model or the person encoder, it tends to overfit to the clothing worn by the target subject. Our finetuning approach successfully retains both the person's identity and the intricate details of the input garments.

\begin{table}[t]
\centering
\scalebox{0.7}{
\begin{tabular}{ |c|cc|  }
\hline
Methods & FID $\downarrow$ & KID $\downarrow$ \\ 
\hline 
Cascaded & $18.523$ & $\mathbf{15.218}$ \\
From Scratch & $21.645$ & $15.781$ \\
Ours & $\mathbf{18.145}$ & $15.227$  \\
\hline
\end{tabular}
}
\caption{\textbf{Quantitative results for ablation studies.} We report FID and KID on our $8,300$ triplets test set.}
\label{table:ablation}
\vspace{-3mm}
\end{table}
\subsection{Single Stage Model vs.\ Cascaded}
\begin{figure*}[htb]
\begin{center}
   \includegraphics[width=1\linewidth]{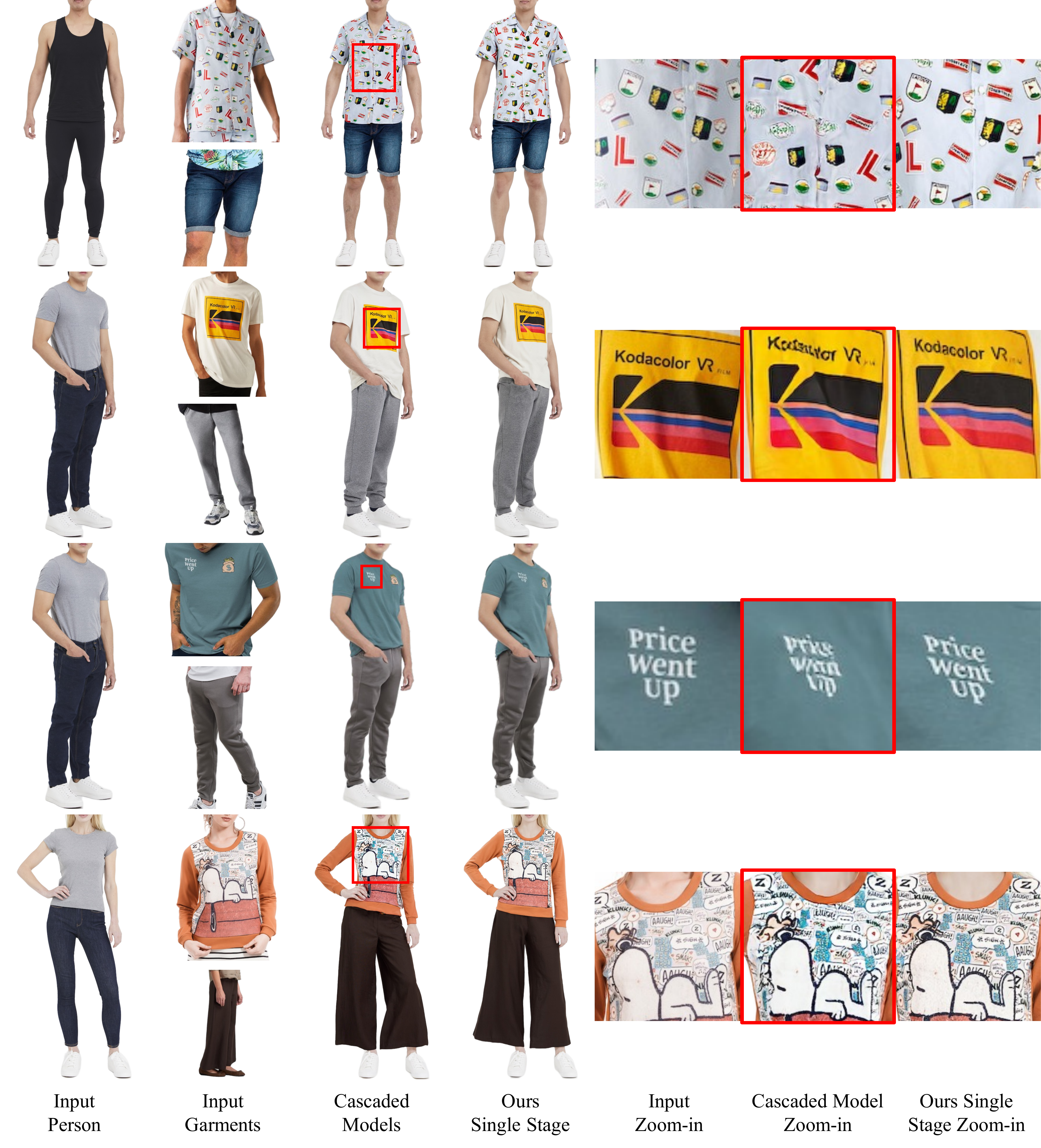}
\end{center}
\vspace{-5mm}
\caption{\textbf{Qualitative comparison for single stage model vs cascaded.} Our proposed single stage model can preserve fine garment details like text and logos under large pose differences. The last three columns visualize zoom-ins of red boxes for input, cascaded variant and single stage model respectively. Please zoom in to see details.}
\label{fig:suppl_ablation_two_stages}
\end{figure*}

Table~\ref{table:ablation} (\nth{1} and \nth{3} rows) presents the FID and KID metrics on our $8,300$ triplets test set, comparing our single-stage model with the cascaded variant. Additionally, Figure~\ref{fig:suppl_ablation_two_stages} offers more qualitative results. While our method does not surpass the cascaded variant in terms of FID and KID scores with significant margin, the qualitative results indicate that it excels at preserving complex garment details, such as texts and logos. This observation aligns with insights from~\cite{podell2023sdxl,kirstain2023pick}, which suggest that FID and KID are more effective at capturing overall visual composition rather than the nuances of fine-grained visual aesthetics.

\subsection{Progressive Training vs.\ Training from Scratch}
\begin{figure*}[htb]
\begin{center}
   \includegraphics[width=1\linewidth]{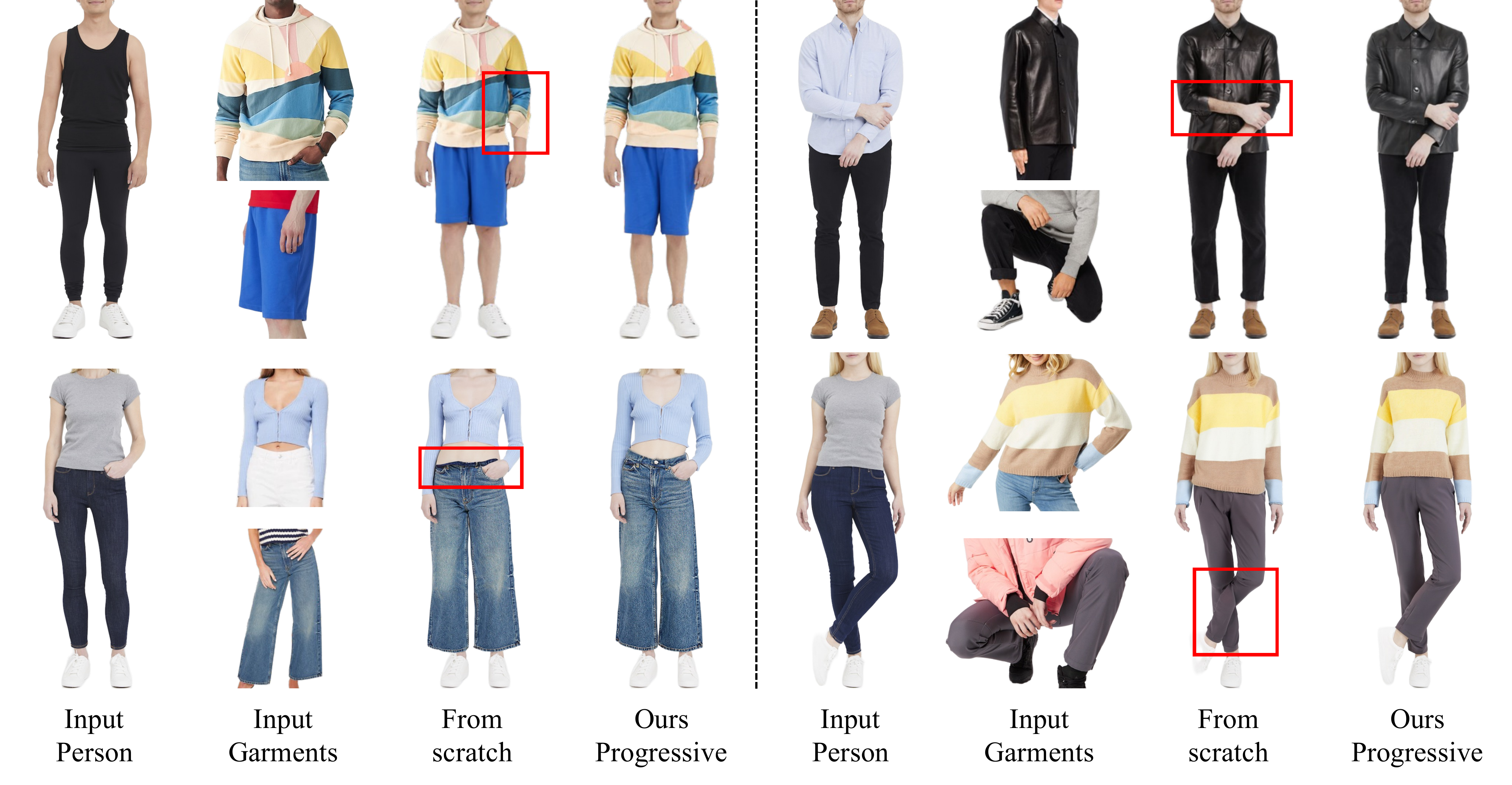}
\end{center}
\vspace{-8mm}
\caption{\textbf{Qualitative comparison for progressive training vs training from scratch.} Training from scratch can not handle complicated garment warping. Red boxes highlight errors of the training from scratch variant. Please zoom in to see details.}
\label{fig:suppl_ablation_from_scratch}
\end{figure*}

Table~\ref{table:ablation} (\nth{2} and \nth{3} rows) reveals that our progressive training strategy yields better results than training from scratch when considering FID and KID scores on our $8,300$ triplets test set. In Figure~\ref{fig:suppl_ablation_from_scratch}, we demonstrate additional qualitative results, suggesting that our progressive training approach is more effective at managing complicated garment warping.

\begin{table}[h]
\centering
\scalebox{0.9}{
\begin{tabular}{ |c|c|c|}
\hline
 & TryOnDiffusion~\cite{Zhu_2023_CVPR_tryondiffusion} & Ours \\ 
\hline 
SSIM $\uparrow$  &  0.883 &  $\mathbf{0.908}$  \\
LPIPS $\downarrow$ &  0.165 & $\mathbf{0.096}$ \\
\hline
\end{tabular}
}
\vspace{-1mm}
\caption{SSIM and LPIPS scores on our $1,000$ paired test data.}
\label{table:paired_data_quantitative}
\vspace{-4mm}
\end{table}
\subsection{Comparison on Paired Test Set}
\begin{figure*}[htb]
\begin{center}
   \includegraphics[width=0.75\linewidth]{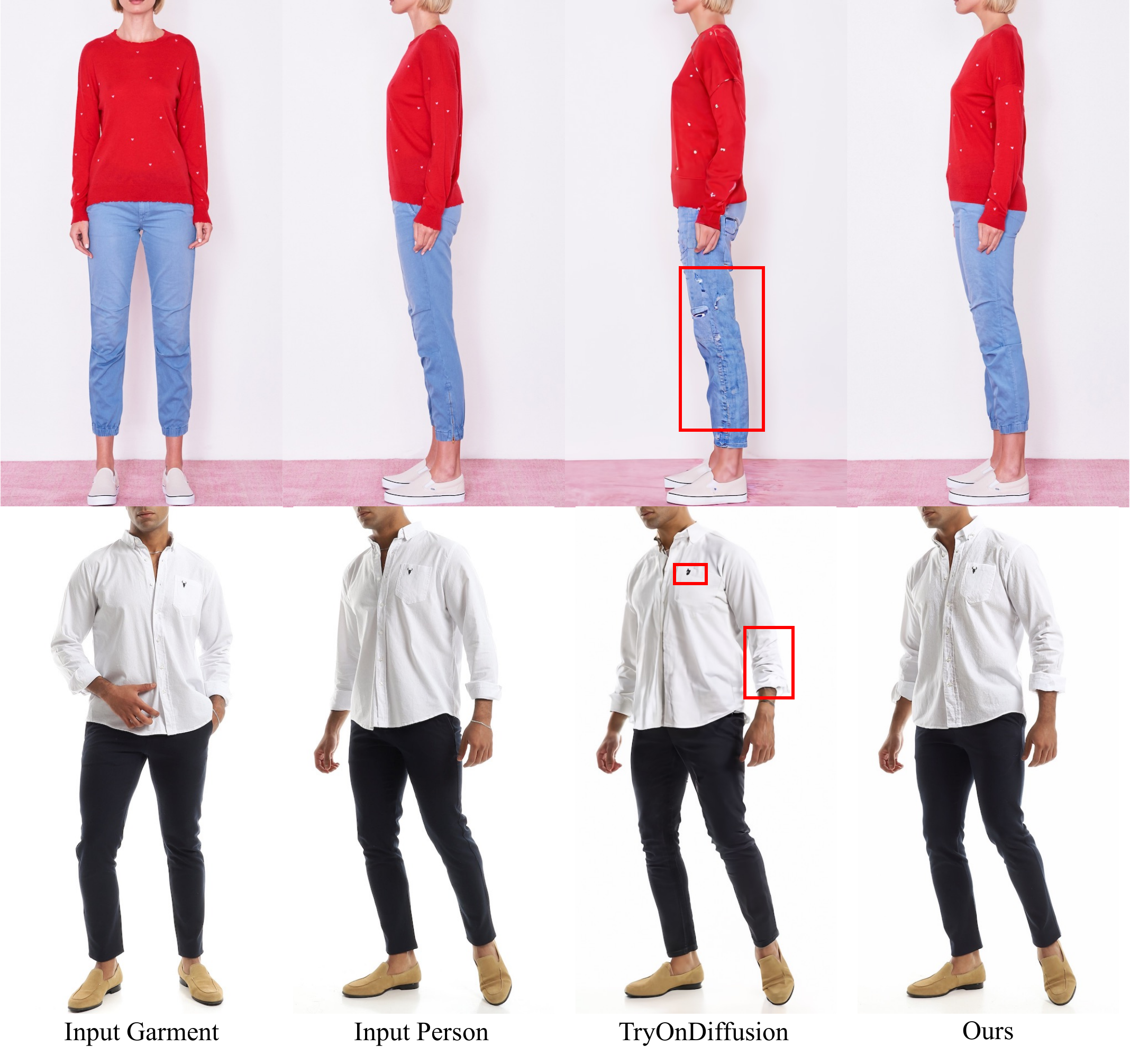}
\end{center}
\vspace{-8mm}
\caption{\textbf{Qualitative comparison on our $1,000$ paired test data.} Red boxes highlight errors of baselines. Zoom in to see details.
}
\label{fig:paired_qualitative}
\vspace{-8mm}
\end{figure*}

We have collected $1,000$ paired test set (not seen during training. Each pair has same person wearing the  garment but under two poses).  Table~\ref{table:paired_data_quantitative} shows that our method achieves better SSIM and LPIPS for the paired data compared to TryOnDiffusion~\cite{Zhu_2023_CVPR_tryondiffusion}. Figure~\ref{fig:paired_qualitative} shows  qualitative results, where our method can better preserve intricate garment details.

\subsection{Additional Qualitative Results}
\begin{figure*}[htb]
\begin{center}
   \includegraphics[width=1\linewidth]{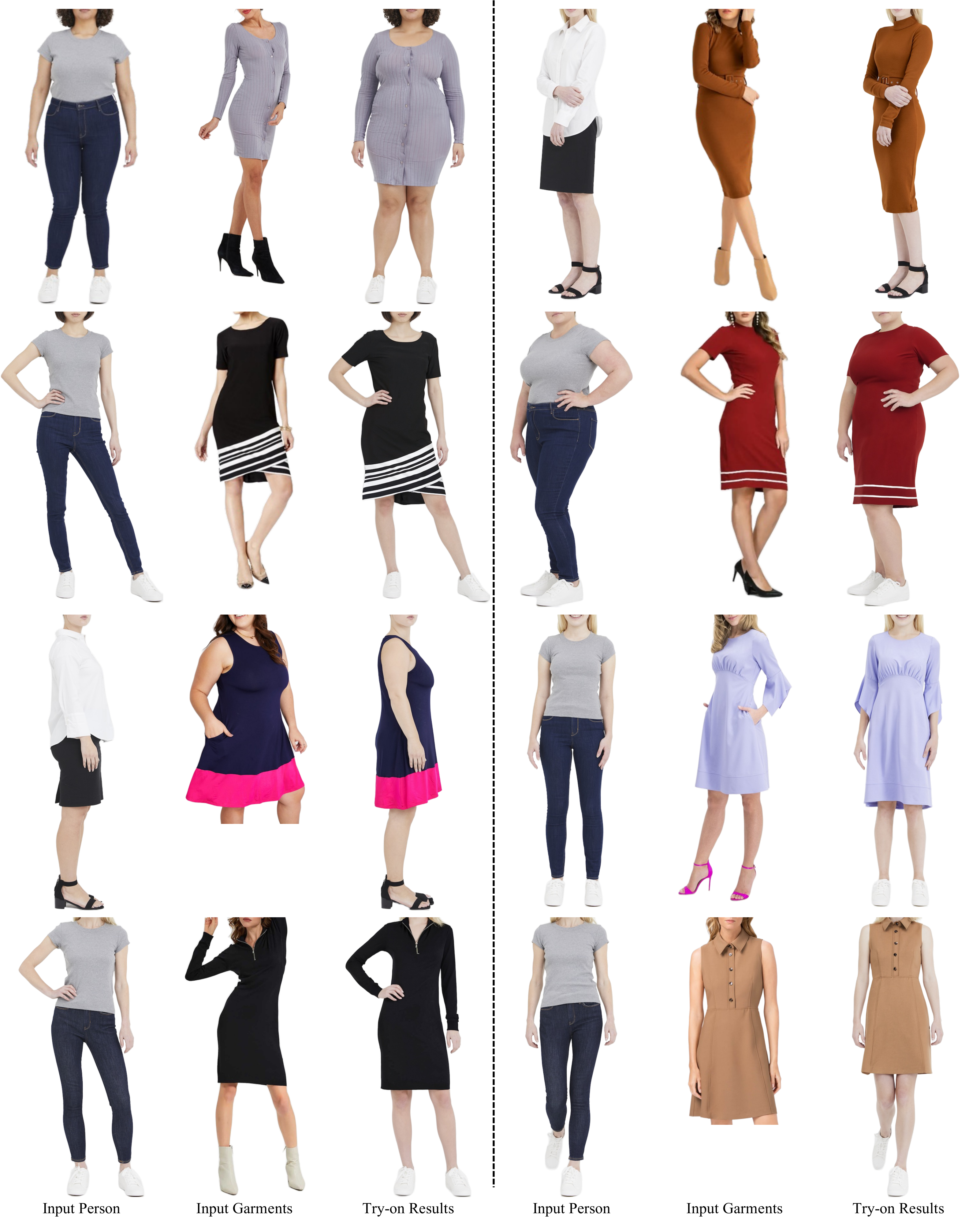}
\end{center}
\vspace{-8mm}
\caption{\textbf{Qualitative results for Dress VTO part one.} Our approach effectively manages complex garment warping and generates realistic wrinkles that align with the person's pose.  Please zoom in to see details.}
\label{fig:suppl_dress_vto1}
\end{figure*}

\begin{figure*}[htb]
\begin{center}
   \includegraphics[width=1\linewidth]{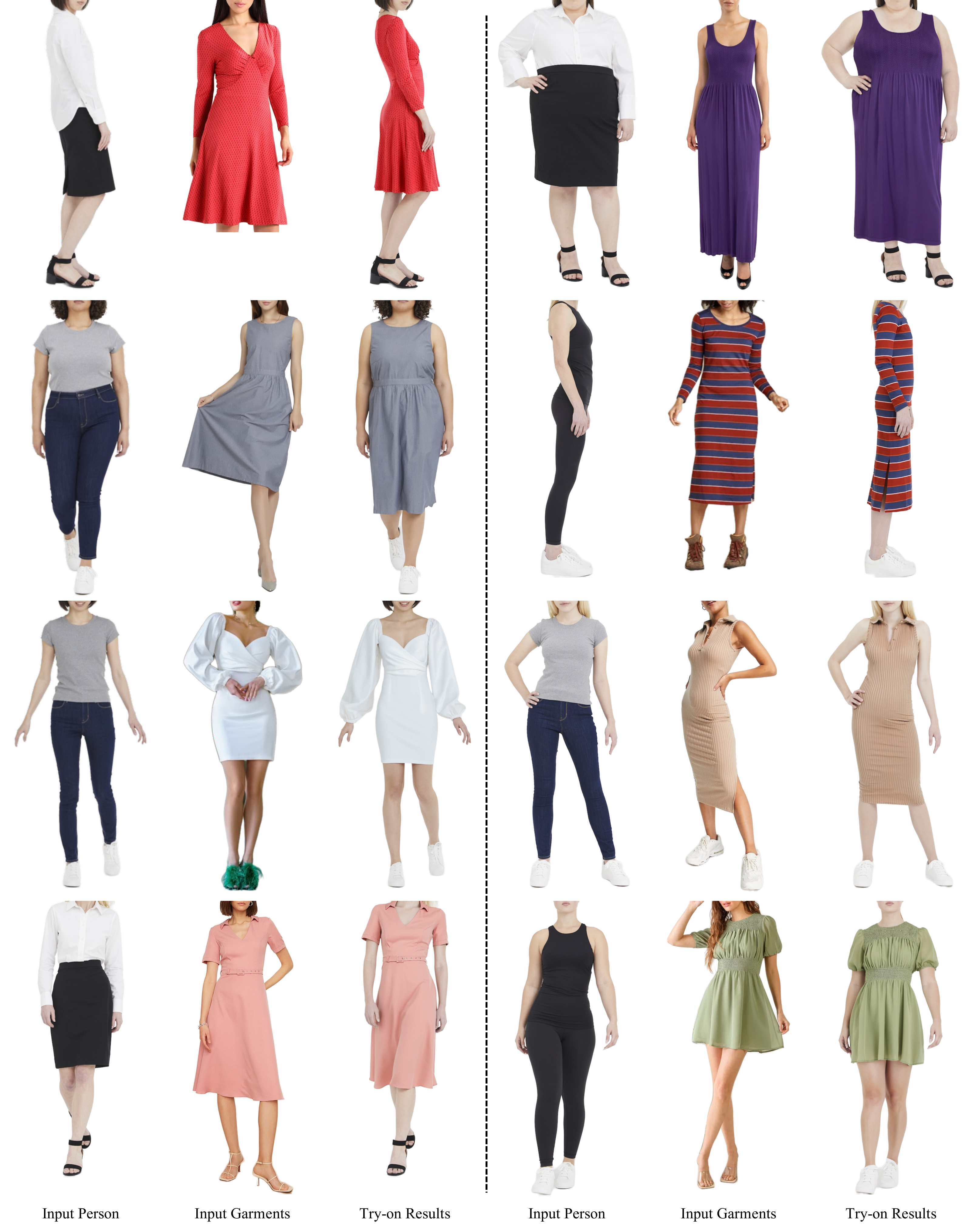}
\end{center}
\vspace{-8mm}
\caption{\textbf{Qualitative results for Dress VTO part two.} Our approach effectively manages complex garment warping and generates realistic wrinkles that align with the person's pose. Please zoom in to see details.}
\label{fig:suppl_dress_vto2}
\end{figure*}

\begin{figure*}[htb]
\begin{center}
   \includegraphics[width=1\linewidth]{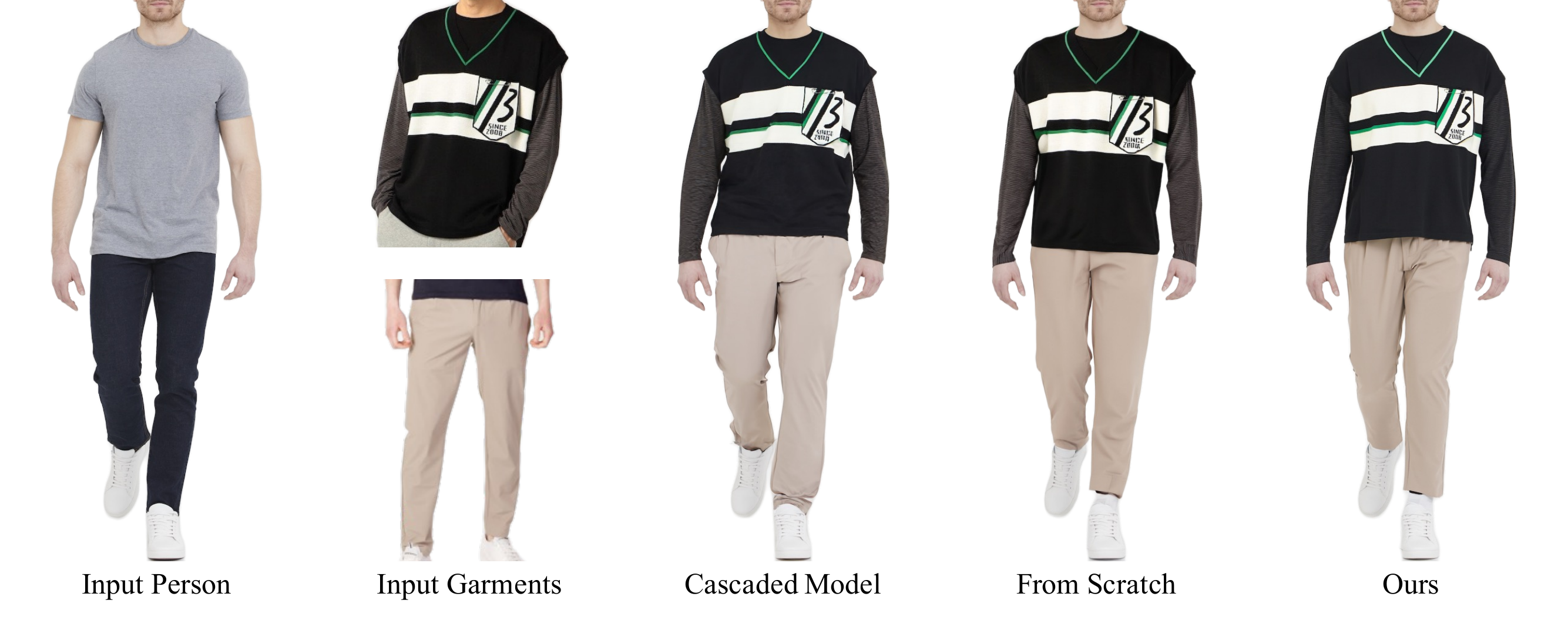}
\end{center}
\vspace{-8mm}
\caption{\textbf{Full images of Figure 6 in the main paper.} Please zoom in to see details.}
\label{fig:suppl_main_fig6_full}
\end{figure*}

\begin{figure*}[htb]
\begin{center}
   \includegraphics[width=1\linewidth]{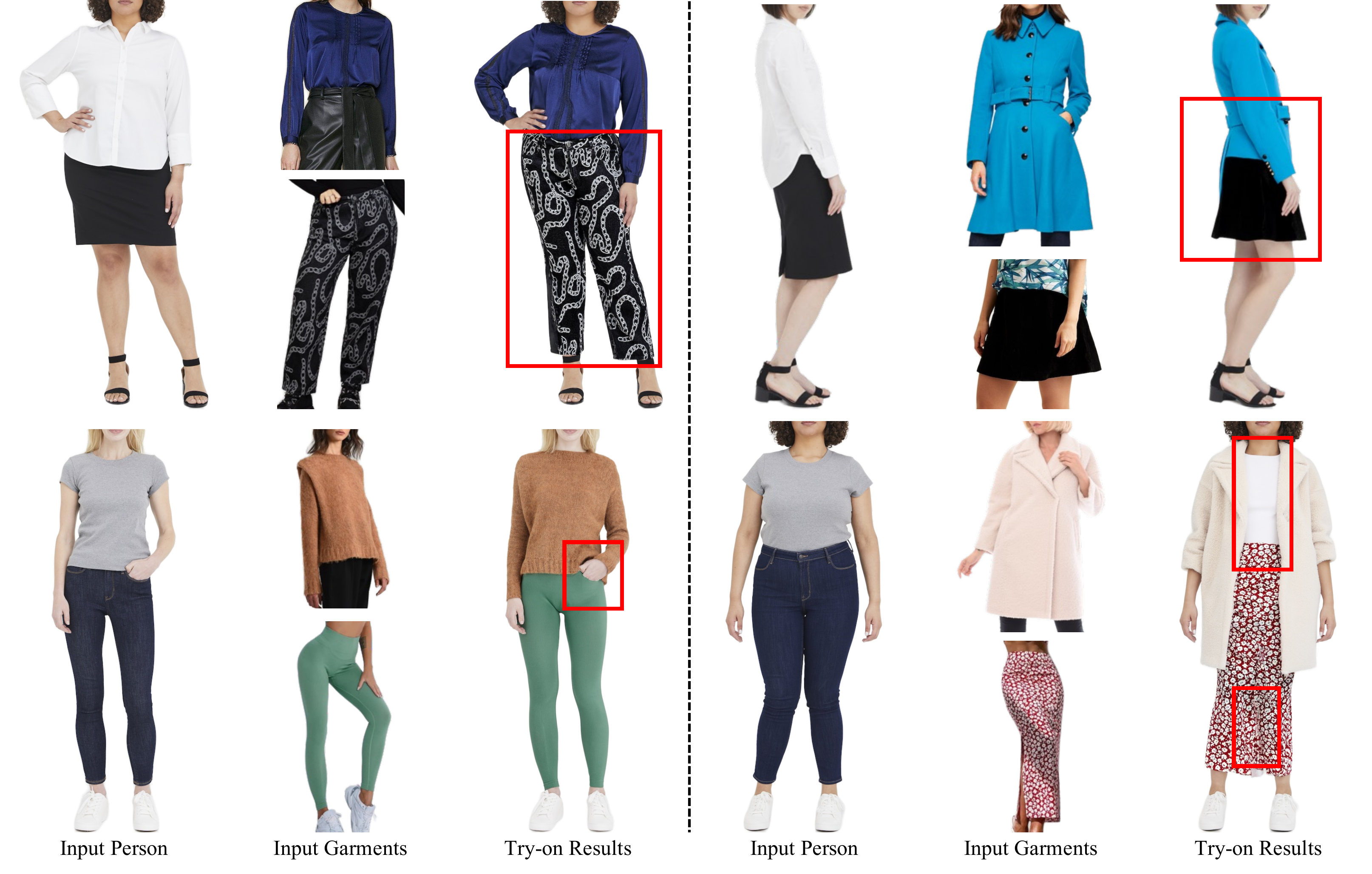}
\end{center}
\vspace{-8mm}
\caption{\textbf{More failure cases.} Top left: our method sometimes suffers from color drift issues for very dark images, which is recognized by diffusion literature~\cite{lin2023common}. Top right: our method fails to generate valid layout for uncommon garment combinations (\eg long coat and skirt). Bottom left: the model attempts to create a pocket to accommodate the occluded left hand. Bottom right: our model could generate a random inner top given ``outer top open'' garment layout.  Additionally, it has difficulties in effectively warping small, densely packed, and irregularly distributed texture patterns.}
\label{fig:suppl_failure_cases}
\end{figure*}

Figure~\ref{fig:suppl_dress_vto1} and ~\ref{fig:suppl_dress_vto2} present try-on results for the dress category (denoted as $I_{g}^{\text{full}}$ in the main paper). Note that our method is able to synthesize realistic folds and wrinkles in dress, well aligned with the person's pose, while preserving the intricate details of the garment. Figure~\ref{fig:suppl_main_fig6_full} visualizes full images of Figure 6 in the main paper. Figure~\ref{fig:suppl_failure_cases} provides more failure cases of our method. Finally, we provide interactive web demos for the mix and match try-on task in the supplementary material.


\end{document}